\documentclass{article} 
\usepackage{iclr2026_conference,times}

\usepackage{amsmath,amsfonts,bm}

\def\eqref#1{equation~\ref{#1}}

\def\1{\bm{1}}

\DeclareMathAlphabet{\mathsfit}{\encodingdefault}{\sfdefault}{m}{sl}
\SetMathAlphabet{\mathsfit}{bold}{\encodingdefault}{\sfdefault}{bx}{n}

\usepackage{mdframed}
\newmdenv[
  linecolor=black,
  linewidth=1pt,
  innerleftmargin=6pt,
  innerrightmargin=6pt,
  innertopmargin=6pt,
  innerbottommargin=6pt,
  skipabove=\baselineskip,
  skipbelow=\baselineskip
]{casestudyframe}

\usepackage{hyperref}
\usepackage{url}
\usepackage{wrapfig}
\usepackage{inconsolata}
\usepackage{multirow}
\usepackage{subcaption}
\usepackage{amsmath}
\usepackage{amssymb}
\usepackage{mathtools}
\usepackage{amsthm}
\usepackage[capitalize,noabbrev]{cleveref}
\usepackage[ruled,vlined,linesnumbered]{algorithm2e}
\theoremstyle{plain}
\newtheorem{theorem}{Theorem}[section]

\theoremstyle{definition}
\newtheorem{definition}[theorem]{Definition}

\theoremstyle{remark}

\usepackage{graphicx}
\usepackage{longtable, booktabs}
\usepackage{listings}
\usepackage{xcolor}

\usepackage[T1]{fontenc}

\lstdefinestyle{mystyle}{
    language=Python,
    backgroundcolor=\color{white},
    commentstyle=\color{gray},
    keywordstyle=\color{blue},
    numberstyle=\tiny\color{gray},
    stringstyle=\color{orange},
    basicstyle=\ttfamily\footnotesize,
    breaklines=true,
    captionpos=b,
    numbers=left,
    numbersep=5pt,
    showspaces=false,
    showstringspaces=false,
    showtabs=false,
    tabsize=4
}

\title{Debugging Tabular Log as Dynamic Graphs}

\author{Chumeng Liang~\thanks{Equal contribution} \\
University of Illinois Urbana-Champaign\\
\texttt{chumengl@illinois.edu}\\
\And
Zhanyang Jin\footnotemark[1]\\
University of Illinois Urbana-Champaign\\
\texttt{zj26@illinois.edu}\\
\And
Zahaib Akhtar\\
Amazon\\
\texttt{akhtz@amazon.com}\\
\And
Mona Pereira\\
Amazon\\
\texttt{monperei@amazon.com}\\
\And
Haofei Yu\\
University of Illinois Urbana-Champaign\\
\texttt{haofeiy2@illinois.edu}\\
\And
Jiaxuan You~\thanks{Corresponding author}\\
University of Illinois Urbana-Champaign\\
\texttt{jiaxuan@illinois.edu}\\
}

\iclrfinalcopy 
\begin{document}

\maketitle

\begin{abstract}
Tabular log abstracts objects and events in the real-world system and reports their updates to reflect the change of the system, where one can detect real-world inconsistencies efficiently by debugging corresponding log entries. However, recent advances in processing text-enriched tabular log data overly depend on large language models (LLMs) and other heavy-load models, thus suffering from limited flexibility and scalability. This paper proposes a new framework, GraphLogDebugger, to debug tabular log based on dynamic graphs. By constructing heterogeneous nodes for objects and events and connecting node-wise edges, the framework recovers the system behind the tabular log as an evolving dynamic graph. With the help of our dynamic graph modeling, a simple dynamic Graph Neural Network (GNN) is representative enough to outperform LLMs in debugging tabular log, which is validated by experimental results on real-world log datasets of computer systems and academic papers.
\end{abstract}

\section{Introduction}
\label{sec:1}
Tabular log data plays a crucial role in representing and tracking the state and evolution of real-world systems. These logs are structured as rows of log entries, each capturing an event involving certain objects and their attributes at a specific time point. Common examples include system logs recording computing services~\citep{zhu2023loghub}, research logs tracking scientific publication activities~\citep{clement2019use}, and interaction logs from multi-agent systems powered by large language models (LLMs)~\citep{zhang2025agent}. Debugging of tabular logs is essential: it allows practitioners to detect anomalies in the original systems through efficient inspection of associated log records. 

Log anomaly detection~\citep{he2016experience} has therefore been a long-standing research field in different niche areas, where data distributions are invariant or have little change. Existing frameworks~\citep{du2017deeplog,meng2019loganomaly,zhang2019logrobust,pei2020subgraph,guo2021logbert,chen2022antibenford} benefit from manually defined data structures or templates for log parsing which are often tailored to certain domains and thus yield absolute success in specific areas like computer system log or financial event log. However, due to this domain-specific principle, designing a general-purpose log debugger always remains challenging. 

\begin{figure*}[t]
\vspace{-0.2cm}
  \includegraphics[width=\textwidth]{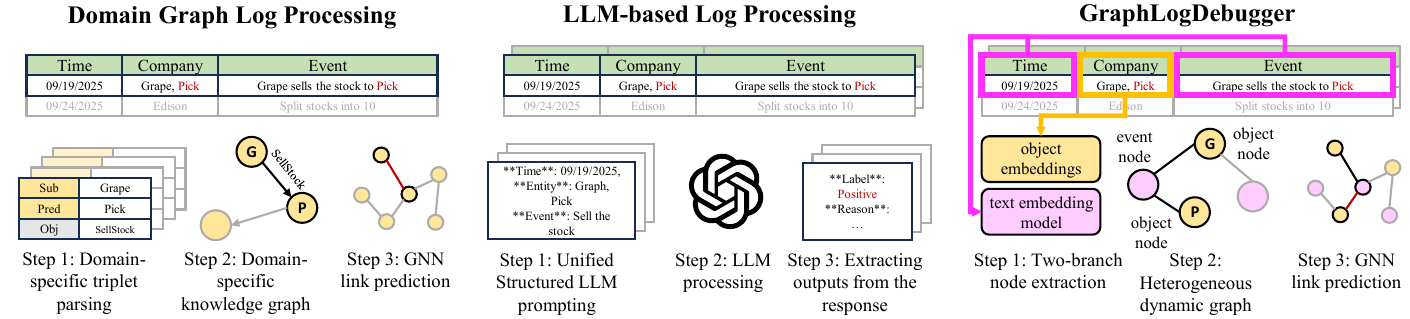}
  \caption{\textbf{Comparing GraphLogDebugger with two existing lines of works}. Processing log with domain-specific graphs requires custom text parsing, which lacks flexibility. LLM-based log processing overcomes this shortcoming by the general comprehension skills of LLMs, but suffers from poor efficiency. GraphLogDebugger combines the advantages of graph representation and those of LLMs and balances well generalizability and scalability.}
  \label{fig:demo}
  \vspace{-0.3cm}
\end{figure*}

Efforts to overcome this challenge have led to two main lines of work, as shown in Figure~\ref{fig:demo}. One stream focuses on graph modeling of the log data~\citep{cheng2020knowledge,zehra2021financial,pang2025guard}, where information in log is gathered in a unified data structure: the graph, such as constructing knowledge graphs or text-rich dynamic graphs for computer system log~\citep{sui2023logkg,li2023glad}. Although these methods are both efficient and powerful, many of them lack flexibility: they still customize static graph structures for certain domains. Another stream explores LLM-based solutions, such as LLM prompting~\citep{yu2023temporal,qi2023loggpt,park2024enhancing} or retrieval-augmented generation (RAG)~\citep{pan2024raglog,zhang2025xraglog,wang2025financial} pipelines. While these methods demonstrate general capabilities in text-based reasoning, thus showing potential of generalization, they often come with significant drawbacks: high computational costs, slow inference, and difficulty scaling to long log streams or resource-constrained settings. 

Inspired by the idea to unify multimodal information in dynamic graphs~\citep{feng2025graph}, we propose \textbf{GraphLogDebugger}, a general and efficient framework for debugging tabular logs through dynamic graph modeling. Our core idea is to interpret tabular log entries as the evolving state of a hidden system, which can be reconstructed as a dynamic heterogeneous graph. We treat objects and events as different types of nodes with text embeddings empowered by modern language embedding models, and use the tabular structure to generate time-stamped connections between them. As new log entries arrive, they incrementally update the dynamic graph, capturing both structural and temporal dependencies. This formulation allows us to apply a lightweight dynamic Graph Neural Network (GNN) to perform online anomaly detection by evaluating the likelihood of new connections. Our approach avoids reliance on heavy LLMs while still capturing rich semantic and relational information in the data. Experimental results on real-world datasets from computer system logs and scientific publication logs validate the effectiveness of our approach. Despite its simplicity, our dynamic GNN framework outperforms LLM-based baselines in both accuracy and efficiency, demonstrating that dynamic graph modeling is a highly expressive yet lightweight alternative. Our contributions can be summarized as follows:
\begin{itemize}
    \item We introduce a novel view of tabular logs as dynamic heterogeneous graphs, bridging the gap between structured attributes and semantic reasoning, and redefine the framework of online log anomaly detection, where object-event connections in each incoming log are evaluated through link prediction on the evolving graph.
    \item We propose a lightweight GNN-based debugger that can efficiently and accurately detect anomalies without using LLMs, and validate its performance on real-world datasets with diverse modalities.
\end{itemize}

\section{Related Works}

\paragraph{Tabular Log Processing.} 
Many real-world logs include structured, time-stamped tabular attributes alongside annotated text fields. Examples come from financial prices paired with event series~\citep{tetlock2007giving,ruiz2012correlating,dong2024fnspid}, scientific publication metadata~\citep{clement2019use,kinney2023semantic}, healthcare records~\citep{johnson2023mimic}, and computer~\citep{zhu2023loghub} and multi-agent system reports~\citep{zhang2025agent}. A key challenge in processing tabular logs lies in capturing multi-attribute correlations while maintaining comprehension of their semantics~\citep{wu2025tabular}. One common approach integrates main attributes recognized by human priors into structured data~\citep{yang2018dcfee,zhao2022utilizing,koval2024financial}, and then subsequently augments the representation by retrieval~\citep{kurisinkel2024text2timeseries,xiao2025enhancing}.  This modeling achieves good performance in domain-specific data, but lacks flexibility and generalizability for adaptation to other fields~\citep{gardner2024large}.

An emerging alternative leverages Large Language Models (LLMs)~\citep{brown2020language}, which have demonstrated strong generalizability in understanding, predicting, and generating tabular data~\citep{liu2023rethinking,zhang2024tablellm,fang2024large,wang2024chain}. By parsing diverse logs into a unified format with LLMs~\citep{zhong2024logparser}, these models can be applied to downstream tasks that require reasoning capabilities, such as predicting stock prices~\citep{yu2023temporal}, electricity demand~\citep{wang2024news}, and future events~\citep{shi2023language,ye2024mirai}. However, LLM-based approaches often suffer from high overheads, complex deployment, and limited throughput. There remains a strong need for lighter-weight alternatives with comparable performance.

\paragraph{Dynamic Graphs.}

Graph Neural Networks (GNNs) have become a foundational paradigm for learning on graph-structured data~\citep{kipf2016semi,hamilton2017inductive,gilmer2017neural}. Static GNN models have benefited from advances in message passing~\citep{battaglia2018relational}, architectural depth~\citep{li2021deepgcns,dwivedi2020benchmarking}, and inductive scalability~\citep{hamilton2017inductive}. However, many real-world systems are dynamic, motivating models that capture both structural and temporal dependencies. Early approaches used recurrent layers or time-aware embeddings~\citep{li2017diffusion,seo2018structured} to extend static GNNs to dynamic settings~\citep{pareja2020evolvegcn,sankar2020dysat,kumar2019predicting}. Recent methods have embraced memory modules~\citep{rossi2020temporal} and temporal encoding~\citep{xu2020inductive} for finer-grained modeling of time-stamped interactions. Building on this trajectory, ROLAND~\citep{you2022roland} offers a framework that adapts static GNNs to dynamic graphs via hierarchical state propagation and live-update evaluation, which inspires numerous new advances in benchmarks~\citep{longa2023temporal,huang2023tgb,zhang2024dtgb}, architectures~\citep{zhu2023wingnn}, explainability~\citep{chen2023tempme}, fairness~\citep{song2023fair}, and robustness~\citep{zhang2023spectral}.

\paragraph{Log Anomaly Detection.}

Log-based anomaly detection has long been a critical task for system reliability, and early neural approaches typically rely on sequence modeling via LSTMs~\citep{du2017deeplog}, CNNs~\citep{lu2018cnn}, and autoencoders~\citep{zhang2021logattn,castillo2022autolog,zhang2023vae}. Others incorporate adversarial training~\citep{duan2021gan,he2023gan}, or temporal networks~\citep{zhang2019logrobust,yang2021plelog}. More recently, pretrained language models have been adopted for log anomaly detection, either via fine-tuning~\citep{guo2021logbert,lee2023lanobert} or prompt-based pipelines~\citep{qi2023loggpt,liu2024logprompt}. Retrieval-augmented~\citep{no2024rapid,pan2024raglog,zhang2025xraglog} methods have further pushed semantic understanding in LLM-based methods. As mentioned, while machine learning-based methods are highly domain-specific, LLM-based methods show some generalizability at a high cost.

One potential solution towards general and scalable methods for log debugging is to introduce dynamic graphs, where we consider tabular log as an evolving system and maintain its digital twin by storing information and relations in a dynamic graph. Early exploration makes use of knowledge graphs~\citep{hogan2021knowledge} with domain specific parsing to generate triplets~\citep{cheng2020knowledge,zehra2021financial,sui2023logkg}. Recent advances adopt dynamic graphs with text-rich nodes to represent tabular log~\citep{li2023glad,pang2025guard}. Nevertheless, these works are either domain specific or LLM-based, yet not escaping from the dilemma between generalizability and scalability.
\section{Preliminaries}
\label{sec:3.1}
Tabular log is the data modality used to report the update of real-world systems from the perspective of states and relations. It can be formally defined by a time series $X=\{x_{0}, x_{1},x_{2},...,x_{{N-1}}\}$ annotated by a timestamps sequence $t_0<t_1<t_2<...<t_{N-1}$, where each of $x_{n}$ is a log entry that contains different attributes $x_{n}^m$ in the table: $x_{n}=\{x_{n}^0,x_{n}^1,x_{n}^2,...,x_{n}^{M-1}\}$. Summarizing the general case of tabular log data in finance~\citep{dong2024fnspid}, healthcare~\citep{johnson2023mimic}, academics~\citep{clement2019use}, and other systems~\citep{zhu2023loghub}, we can separate attributes in the tabular log into three types:

\begin{itemize}
    \item\textbf{Object:} Attributes that represent stand-alone objects in the tabular log, such as companies in the financial news log and cities in the medical record log. 
    \item\textbf{Event:} Attributes that describe an event with text, for example, news content in the financial news log and record content in the medical record log. These attributes are usually the center of log entries, where other attributes supplement details and involved objects of the event. Without loss of generality, one log entry only has one Event attribute, because we could merge the text sections of different event attributes into one.
    \item\textbf{Feature:} Attributes that describe features related to the event or objects. For instance, the age is a feature of the patient object in the medical record log. Timestamp $t_n$ is a special type of feature that provides the details about the time of the event. 
\end{itemize}

In practice, we find that Objects and Features are mutually convertible. For example, the address of the company could be either an independent object or a feature of the company object in the financial log. Hence, the arrangement of Objects and Features is a hyperparameter that needs pre-definition.

While tabular log abstracts the change of real-world systems, it is expected that we could detect inconsistencies of the system from the corresponding tabular log. Based on the above categorization, we could then define three types of anomalies and corresponding anomaly detection tasks in the tabular log. Give a log entry $x_{n}=\{x_{n}^0,x_{n}^1,x_{n}^2,...,x_{n}^{M-1}\}$ in the tabular log $X$:
\begin{itemize}
    \item\textbf{Object anomaly:}  Let $\{o_n^0,o_n^1,o_n^2,...,o_n^{P-1}\} (P<M)$ be the object set. We have label $y_n=\{y_n^0,y_n^1,y_n^2,...,y_n^{P-1}\}$, where $y_n^m=0$ means that $o_n^p$ is a normal object and $y_n^m=1$ means that $o_n^p$ is an abnormal object for the log entry $x_{n}$.
    \item\textbf{Event anomaly:} Let $s_n$ be the event. We have a label $y_n$, where $y_n=0$ means that $s_n$ is a normal event and $y_n=1$ means that $s_n$ is an abnormal event for the log entry $x_{n}$.
    \item\textbf{Feature anomaly:} Let $\{f_n^0,f_n^1,f_n^2,...,f_n^{Q-1}\} (Q<M)$ be the feature set. We have label $y_n=\{y_n^0,y_n^1,y_n^2,...,y_n^{Q-1}\}$, where $y_n^q=0$ means that $o_n^q$ is a normal feature and $y_n^q=1$ means that $o_n^q$ is an abnormal feature for the log entry $x_{n}$.
\end{itemize}

\begin{wrapfigure}{r}{0.57\textwidth}
\vspace{-0.6cm}
  \includegraphics[width=\linewidth]{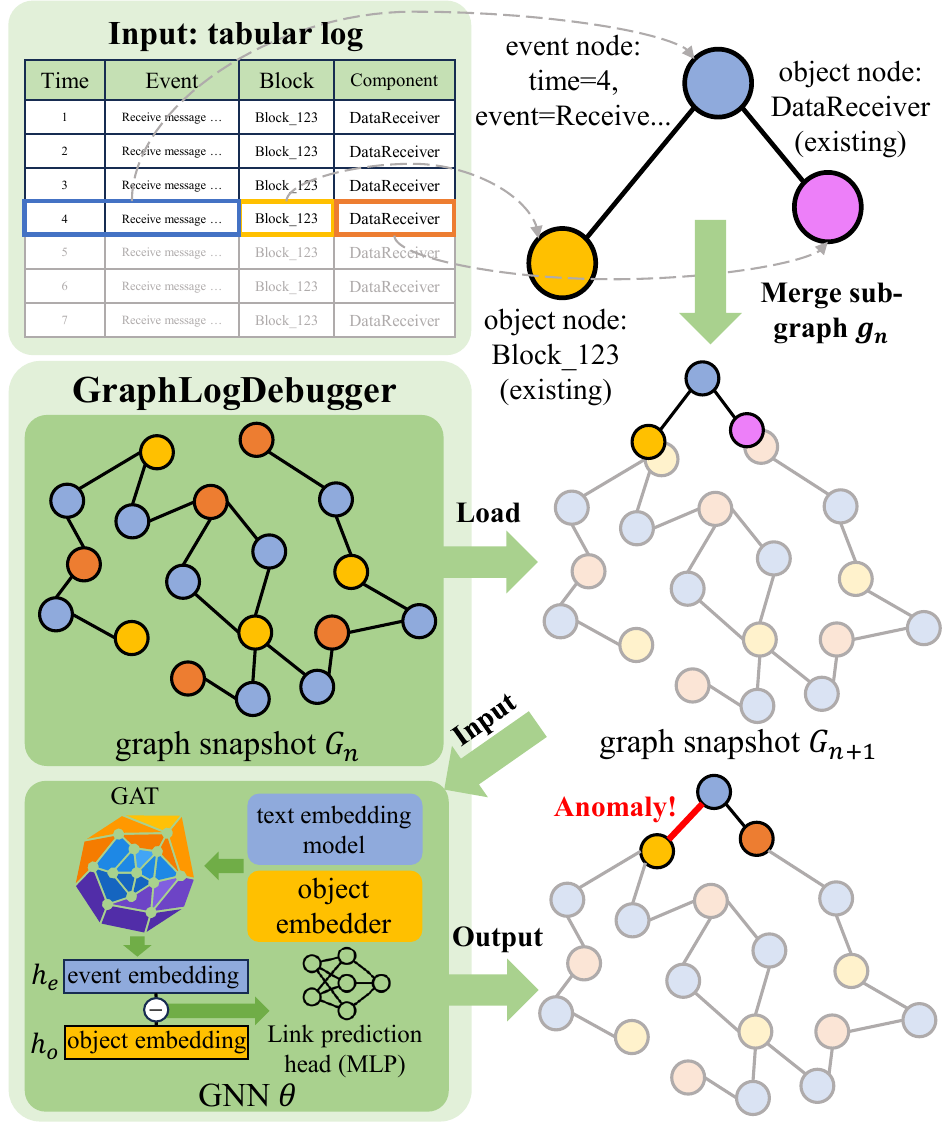}
  \caption{\textbf{GraphLogDebugger framework}. The framework checkpoints the GNN $\theta$ and the dynamic graph snapshot $G_n$. When a new log entry emerges, we first extract a sub-graph $g_n$ and use it to update the graph. Then, we predict the links introduced by $g_n$ in the dynamic graph by GNN $\theta$, whose results indicate the anomaly.}
  \label{fig:main}
  \vspace{-2.5cm}
\end{wrapfigure}

Considering the tabular log $X$ as an online system where new log entries come dynamically in time order, we could then define the anomaly detection in tabular log as an online anomaly detection task:

\begin{definition}[\textbf{Online Anomaly Detection of Tabular Log}]
\label{def:1}
Given an online system that dynamically produces log entry $x_{n}$, online anomaly detection for tabular log predicts its anomaly label $y_n$ based on the historical log entries $X_{n}=\{x_0, x_1,x_2,...,x_{n-1}\}$ 
\end{definition}

Notably, object anomaly and feature anomaly are isomorphic. Considering the fact that objects and features are convertible, the rest of this paper only studies object anomaly and event anomaly.

\section{GraphLogDebugger}

We first integrate tabular log to a heterogeneous dynamic graph (Section~\ref{sec:3.2}). Then, we reformulate the online anomaly detection of tabular log as a dynamic graph anomaly detection problem(Section~\ref{sec:3.3}). Finally, we apply a dynamic graph neural network (GNN) to debug the tabular log (Section~\ref{sec:3.4}).
\subsection{Integrating online tabular log to dynamic graphs}
\label{sec:3.2}
Objects and events in the same log entry are naturally connected in the tabular log, from which we could construct graphs. To this end, we first define the graph structure within one log entry. As shown in Figure~\ref{fig:main} (upper section), we consider both unique objects and the event in the new coming log entry as nodes $v$. Between each object-event pair in one log entry, we connect an edge $e$. This yields a sub-graph $g_n$ for each log entry $x_{n}$.

Figure~\ref{fig:main} (middle section) illustrates the composition of a dynamic graph $\mathcal{G}$ that stores the information of all historical log entries. This dynamic graph gathers all sub-graphs of log entries. We merge identical object nodes so that these sub-graphs are connected. Note that every event node should be unique. Every time a new log entry $x_n$ emerges, we construct a sub-graph $g_n$ accordingly and merge it into the dynamic graph $\mathcal{G}$. We denote the snapshot of $\mathcal{G}$ at time point $t_n$ by $G_n$.

\subsection{Debugging tabular log graphs as dynamic graphs}
\label{sec:3.3}
Following the above integration, we transfer the online anomaly detection of tabular log defined in Section~\ref{sec:3.1} into an anomaly detection problem of dynamic graph $\mathcal{G}$~\citep{ekle2024anomaly}:
\begin{itemize}
    \item\textbf{Object anomaly detection:} The object anomaly occurs when the edge $e$ between an object node and an event node is abnormal in the latest sub-graph $g_n$. This anomaly could then be detected by link prediction in the dynamic graph $\mathcal{G}$~\citep{you2022roland}, where a GNN is applied to predict the likelihood of $e$. If the likelihood exceeds a threshold, we consider the edge as normal. Otherwise, we consider the edge as an anomaly. 
    \item\textbf{Event anomaly detection:} The event anomaly occurs when an abnormal event $s$ is placed in the wrong entry in the latest sub-graph $g_n$. This means that all edges between this event is anomalies. We can therefore apply link prediction to all edges in the sub-graph and threshold the overall predicted likelihoods to determine the anomaly label of the event.
\end{itemize}
In a nutshell, the goal of our anomaly detection in Section~\ref{sec:3.1} is equivalent to predicting the likelihood of all edges in the latest log entry sub-graph $g_n$, based on the dynamic graph snapshot $G_n$ that stores all historical log entries of tabular log $X_n$. We finally transfer the online anomaly detection of tabular log into an anomaly detection problem in dynamic graphs, with notations omitted to Table~\ref{tab:notation}:
\begin{definition}[\textbf{Online Anomaly Detection of Tabular Log (Dynamic Graph)}]
\label{def:2}
Given the snapshot $G_n$ of a dynamic graph $\mathcal{G}=\{G_n\}_{n=0}^{N-1}$ and the new coming sub-graph $g_n=G_{n+1}\setminus G_n$, the goal is to predict the label $y_n$ of links in $g_n$.
\end{definition}

\subsection{Designing the GNN for dynamic graph anomaly detection}
\label{sec:3.4}
Figure~\ref{fig:main} (bottom section) demonstrates the basic process of using our GNN to predict link anomaly labels. For details, our GNN $\theta$ takes the dynamic graph snapshot $G_n$ and the incoming sub-graph $g_n$ as inputs and predicts the likelihood of all links in $g_n$. The GNN consists of three parts: the node embedder, the GNN backbone, and the prediction head. First, the node embedder offers heterogeneous embeddings for all objects in $G_n$ and $g_n$ and event nodes in $G_n$. We exclude new events in $g_n$ because we do not expect the outputs of the model will be interfered with by the graph structure in $g_n$. We assign a unique learnable embedding for each object, and use a pre-trained text embedding model to embed existing events. We also concatenate a time embedding to the event embedding based on the coming time $t_n$ for event $s$. Our GNN backbone is adapted from the graph attention network (GAT)~\citep{velivckovic2017graph}, where we use two separate MLPs to map objects and events to the same space and apply GAT layers for message passing. The prediction head predicts the link between all object-event pairs in $g_n$. We first compare object embeddings after GAT layers and event text embeddings by reduction. We then pass the result to an MLP with Sigmoid activation to get the likelihood. We omit more details in the design space of the model in Appendix~\ref{appendix:model}.

Our GNN is trained under the setting of unsupervised anomaly detection~\citep{pang2021deep}: We separate the dataset into a training split and a test split by chronological order. In the training stage, all links in $g_n$ are normal, and we provide negative examples for training by randomly sampling object-event pairs that are not connected. We append these fraud links to the ground-truth links to balance the label distribution and use them to train the GNN. After backpropagation, we finally update the dynamic graph snapshot $G_n$ with subgraph $g_n$. Alg~\ref{alg:1} summarizes the training algorithm. 

In the test stage, we first resume the GNN as well as the latest dynamic graph snapshot $G_n$. This achieves the warm start of our debugger system. Then, we construct sub-graphs from the coming log entries and take them as parts of the evolving dynamic graph $\mathcal{G}$ that we succeed from the pre-trained GNN. The evaluation process is summarized in Alg~\ref{alg:2}.

\begin{figure*}[tbp]
\vspace{-0.5cm}
  \centering
  \begin{minipage}[t]{0.485\textwidth}
    \vspace{0pt}
    \begin{algorithm}[H]
      \small
      \caption{GraphLogDebugger: Online training for dynamic-graph anomaly detection}\label{alg:1}
      \DontPrintSemicolon
      \SetKwInOut{Input}{Input}\SetKwInOut{Output}{Output}
      \Input{Training log$X_{\text{train}}=\{x_{t_0},\dots,x_{t_{K-1}}\}$; GNN $\theta$; text embedding model $\mathcal{F}$; negative sampling ratio $\rho$; threshold $\tau$; }
      \Output{Trained parameters $\theta^\star$, dynamic graph snapshot $G_{t_{K}}$}
      Initialize $\mathcal{G}:\mathcal{V}=\emptyset,\mathcal{E}=\emptyset$\;
      \For{$k=0,\dots,K-1$}{
      \tcp{\textbf{Integrate the incoming log entry}}
        Build sub-graph $g_k$: creating nodes for object set $\mathcal{V}_k^o$ and event set $\mathcal{V}_k^e$ in $x_{t_k}$ and connecting object-event pairs in $x_{t_k}$ with edges\;
        $\mathcal{G}\leftarrow (\mathcal{V}\cup \mathcal{V}_k^o,\mathcal{E})$\;
        Positive set $\mathcal{E}^+_k \leftarrow$ object-event links in $g_k$.\;
        Negatives $\mathcal{E}^-_k$ by drawing $\rho\cdot|\mathcal{E}^+_k|$ non-existent object--event pairs in $\mathcal{G}$.\;
        \BlankLine
        \tcp{\textbf{Embed nodes}}
        Compute object embeddings $h_o$ with GNN $\theta$: $h_o=f_\theta(\mathcal{G})$\;
        Compute new event embeddings $h_e$ with text embedding model $\mathcal{F}$: $h_e=\mathcal{F}(\mathcal{V}_k^e)$\;
        \BlankLine
        \tcp{\textbf{Predict links \& Compute the loss}}
        For each pair $(o,e)\in \mathcal{E}^+_k\cup\mathcal{E}^-_k$, compute score
        $s_{o,e}=\sigma(\text{MLP}(\text{reduce}(h_o,\,h_e)))$.\;
        Compute balanced BCE loss $\mathcal{L}_k$ on labels (1 for $\mathcal{E}^+_k$, 0 for $\mathcal{E}^-_k$) and update $\theta\leftarrow\theta-\eta\nabla_\theta\mathcal{L}_k$.\;
        \BlankLine
        \tcp{\textbf{Updating the dynamic graph}}
        $\mathcal{G}\leftarrow (\mathcal{V}\cup \mathcal{V}_k^e,\mathcal{E})$\;
      }
        Repeat Step 1-10 for epochs\;
      \Return{$\theta$}\;
    \end{algorithm}
  \end{minipage}
  \hfill
  \begin{minipage}[t]{0.485\textwidth}
    \vspace{0pt}
\begin{algorithm}[H]
  \small
  \caption{GraphLogDebugger: Online evaluation for dynamic-graph anomaly detection}\label{alg:2}
  \DontPrintSemicolon
  \SetKwInOut{Input}{Input}\SetKwInOut{Output}{Output}
  \Input{Test log $X_{\text{test}}=\{x_{t_K},\dots,x_{t_{N-1}}\}$; trained GNN $\theta^\star$; last snapshot $G_{t_K}$; text embedding model $\mathcal{F}$; threshold $\tau$}
  \Output{Per-time link predictions $\{\mathcal{R}_{t_n}\}_{n=K}^{N-1}$; updated snapshot $G_{t_N}$}
  Initialize $\mathcal{G}\leftarrow G_{t_K}$ and load $\theta^\star$\;
  \For{$n=K,\dots,N-1$}{
    \tcp{\textbf{Integrate the incoming log entry}}
    Build sub-graph $g_n$ from $x_{n}$ with object set $\mathcal{V}_n^o$, new event set $\mathcal{V}_n^e$, and observed links $\mathcal{E}^+_n$\;
    $\mathcal{G}\leftarrow (\mathcal{V}\cup \mathcal{V}_k^o,\mathcal{E})$\;
    \BlankLine
    \tcp{\textbf{Embed nodes}}
    Compute object embeddings on current snapshot: $h_o=f_{\theta^\star}(\mathcal{G})$\;
    Compute embeddings for new events (time-aware): $h_e=\mathcal{F}(\mathcal{V}_n^e)$\;
    \BlankLine
    \tcp{\textbf{Predict links}}
    For each $(o,e)\in\mathcal{E}^+_n$, compute
    $s_{o,e}=\sigma(\text{MLP}(\text{reduce}(h_o,\,h_e)))$ and set
    $\hat{\ell}_{o,e}=\mathbb{1}[s_{o,e}\ge \tau]$\;
    \textbf{Link prediction results: }$\mathcal{R}_{t_n}\leftarrow\{(o,e,s_{o,e},\hat{\ell}_{o,e})\mid (o,e)\in\mathcal{E}^+_n\}$\;
    \BlankLine
    \tcp{\textbf{Update the dynamic graph}}
    \textcolor{orange}{Accepted links $\hat{\mathcal{E}}^+_n=\{(o,e)\in\mathcal{E}^+_n\mid \hat{\ell}_{o,e}=1\}$}\;
    \textcolor{orange}{Accepted new events $\hat{\mathcal{V}}^e_n=\{e\in\mathcal{V}_n^e\mid \exists o:(o,e)\in\hat{\mathcal{E}}^+_n\}$.}\;
    \textcolor{orange}{$\mathcal{G}\leftarrow (\ \mathcal{V}\cup \hat{\mathcal{V}}^e_n,\ \mathcal{E}\cup \hat{\mathcal{E}}^+_n\ )$\;}
  }
  \Return{$\{\mathcal{R}_{t_n}\}_{n=K}^{N-1}$ and $G_{t_N}=\mathcal{G}$}\;
\end{algorithm}
  \end{minipage}
\vspace{-0.5cm}
\end{figure*}


\section{Experiments}
\label{sec:5}
\subsection{Experimental settings}
\textbf{Datasets.} Our work provides a general framework of debugging different types of tabular log under the online setting. To validate this point, we span our experiments over datasets covering four different fields: (1) \textbf{Arxiv}: Tabular log recording the timestamps (from 2007-2025), the title and the authors of machine learning papers from the Arxiv~\citep{clement2019use} API; (2) \textbf{HDFS}: system log of Hadoop Distributed File System designed to run on commodity hardware~\citep{xu2009detecting,zhu2023loghub}, including the event content together with the objects related to the event; (3) \textbf{Analyst}: the commentary records on the finance by analysts, including title, author, and other features of posts~\footnote{\url{www.kaggle.com/datasets/miguelaenlle/massive-stock-news-analysis-db-for-nlpbacktests}}; (4) \textbf{Landslide}: event catalog reporting the global landslide~\footnote{\url{https://catalog.data.gov/dataset/global-landslide-catalog-export}}. These datasets contain both text-rich attributes and categorical attributes with diversified semantics, thus being challenging to process in one framework efficiently. Table~\ref{tab:dataset-summary} demonstrates the basic statistics of our four datasets. 
\begin{table*}[t]
\centering
\caption{\textbf{Statistics and details of the four datasets for tabular log debugging.}}
\vspace{-0.2cm}
\label{tab:dataset-summary}
\centering
\resizebox{\linewidth}{!}{
\begin{tabular}{|lcccccc|}
\toprule
\textbf{Dataset} & \textbf{Domain} & \textbf{\#Entries} & \textbf{\#Objects} & \textbf{Event Attr.} & \textbf{Obj Attr.} & \textbf{Anomaly Type} \\
\midrule
Arxiv  & Sci. Pub. & 20,000 & 17316 & title & authors & Event/Object\\
HDFS & System Log & 20,000 & 2150 & Content & Component,EventId,BlockId & Object\\
Analyst & Finance & 20,000 & 3901 & headline & publisher & Event\\
Landslide & Geology & 20,000 & 8565 & description  & title, category,trigger,country  & Object\\
\bottomrule
\end{tabular}
}
\vspace{-0.5cm}
\end
{table*}

 We limit the maximum length of all tabular logs to 20,000 by slicing the original datasets. This is because LLM-based baseline methods are costly and not scalable, as discussed in the introduction. To ensure randomness, we randomly pick slices with a length of 20,000 from the whole sliced dataset. For datasets with multiple object attributes, we evaluate object anomaly detection, while for those with only one object attribute, we evaluate event anomaly detection, where event and object anomaly detection are equivalent. We summarize the basic setting of our four datasets in Table~\ref{tab:dataset-summary}.  
\begin{table*}[t]
  \centering
  \setlength\tabcolsep{3pt}
  \vspace{-5mm}
 \caption{\textbf{Our proposed GraphLogDebugger outperforms representative baselines on detection effectiveness and efficiency across diverse methods and datasets.} Higher is better for detection effectiveness; lower GFLOPs and higher Throughput are preferred for efficiency. ``*'' suggests some baselines always predict non-anomaly cases, leading to a 0 prediction, recall, and F1 score.}
  \label{tab:res-detection-efficiency}
  \resizebox{\textwidth}{!}{
  \begin{tabular}{lcccccc}
    \toprule
    & \multicolumn{6}{c}{Dataset: \textbf{Arxiv} \hspace{3mm} Task: \textbf{Event Anomaly}} \\
    \multirow{2}{*}{Method} & \multicolumn{4}{c}{Detection Effectiveness} & \multicolumn{2}{c}{Efficiency} \\
    \cmidrule(lr){2-5} \cmidrule(lr){6-7}
               & Acc. & Prec. & Recall & F1 & GFLOPs & Throughput (it/s) \\
    \midrule
    MLP &0.570 ± 0.205 &0.556 ± 0.189 &0.893 ± 0.331 &
0.676 ± 0.028 &  11.35   &  825.0 ±1429.0 \\
    RAG (Llama3-70b,$k$=5) &0.408 ± 0.038  &0.426 ± 0.026 &0.527 ± 0.029 &0.471 ± 0.019 &$\sim10^5$  & 0.204 ± 0.009 \\
    RAG (GPT-oss-20b,$k$=5) &0.770 ± 0.106&0.771 ± 0.157 &0.773 ± 0.014 &0.772 ± 0.085 &$\sim10^4$  & 0.145 ± 0.018 \\
    RAG (Llama3-70b,$k$=10) &0.377 ± 0.014 &0.400 ± 0.004 &0.493 ± 0.038 &0.442 ± 0.014 &$\sim10^5$  & 0.204 ± 0.015 \\

    RAG (GPT-oss-20b,$k$=10) &0.803 ± 0.090 &0.798 ± 0.099 &0.813 ± 0.087 &0.805 ± 0.087 &$\sim10^4$  & 0.149 ± 0.014 \\
    \midrule
    GraphLogDebugger (Ours) & \textbf{0.957} ± 0.040 &\textbf{0.920} ± 0.069 &\textbf{1.000} ± 0.000 &\textbf{0.959} ± 0.037 &40.39  & 627.662 ± 7.623 \\
    \bottomrule
  \end{tabular}}

  \vspace{0.2em}

  \resizebox{\textwidth}{!}{
  \begin{tabular}{lcccccc}
    & \multicolumn{6}{c}{Dataset: \textbf{Arxiv} \hspace{3mm} Task: \textbf{Object Anomaly}} \\
    \multirow{2}{*}{Method} & \multicolumn{4}{c}{Detection Effectiveness} & \multicolumn{2}{c}{Efficiency} \\
    \cmidrule(lr){2-5} \cmidrule(lr){6-7}
               & Acc. & Prec. & Recall & F1 & GFLOPs & Throughput (it/s) \\
    \midrule
    MLP &0.570 ± 0.000 &0.538 ± 0.000 &0.990 ± 0.000 &0.697 ± 0.000 &  11.35  &552.0 ± 167.0  \\
    RAG (Llama3-70b,$k$=5) &0.455 ± 0.033 &0.468 ± 0.026 &0.667 ± 0.100 &0.550 ± 0.052 &$\sim10^5$  & 0.208 ± 0.018 \\
    RAG (GPT-oss-20b,$k$=5) &0.597 ± 0.019 &0.564 ± 0.016 &0.850 ± 0.025 &0.678 ± 0.004 &$\sim10^4$  & 0.039 ± 0.018 \\
    RAG (Llama3-70b,$k$=10) &0.463 ± 0.019 &0.474 ± 0.011 &0.673 ± 0.052 &0.556 ± 0.012 &$\sim10^5$  & 0.154 ± 0.114 \\
    RAG (GPT-oss-20b,$k$=10) &0.598 ± 0.038 &0.563 ± 0.023 &\textbf{0.880} ± 0.066 &0.687 ± 0.035 &$\sim10^4$  & 0.039 ± 0.004 \\
    \midrule
    GraphLogDebugger (Ours) & \textbf{0.685} ± 0.065 & \textbf{0.637} ±0.082 &0.870 ± 0.099  & \textbf{0.734} ± 0.024 &40.39  & 592.073 ± 16.513 \\
    \bottomrule

  \end{tabular}}

  \vspace
{0.2em}

  \resizebox{\textwidth}{!}{
  \begin{tabular}{lcccccc}
    & \multicolumn{6}{c}{Dataset: \textbf{HDFS} \hspace{3mm} Task: \textbf{Object Anomaly}} \\
    \multirow{2}{*}{Method} & \multicolumn{4}{c}{Detection Effectiveness} & \multicolumn{2}{c}{Efficiency} \\
    \cmidrule(lr){2-5} \cmidrule(lr){6-7}
               & Acc. & Prec. & Recall & F1 & GFLOPs & Throughput (it/s) \\
    \midrule
    MLP &0.799 ± 0.053 &0.801 ± 0.052 &0.989 ± 0.000 &0.885 ± 0.032 &1.4 &479.0 ± 194.0 \\
    RAG (Llama3-70b,$k$=5) &0.165 ± 0.022 &0 ± 0* &0 ± 0* &0 ± 0* &$\sim10^5$  & 0.162 ± 0.085 \\
    RAG (GPT-oss-20b,$k$=5) &0.138 ± 0.029 &0 ± 0* &0 ± 0* &0 ± 0* &$\sim10^4$  & 0.183 ± 0.010 \\
    RAG (Llama3-70b,$k$=10) &0.173 ± 0.040 &0 ± 0* &0 ± 0* &0 ± 0* & $\sim10^5$ & 0.194 ± 0.004 \\
    RAG (GPT-oss-20b,$k$=10) &0.138 ± 0.029 &0 ± 0* &0 ± 0* &0 ± 0* &$\sim10^4$  & 0.192 ± 0.027 \\
    \midrule
    GraphLogDebugger (Ours) &\textbf{0.989} ± 0.023 &\textbf{1.000} ± 0.000 &\textbf{0.987} ± 0.029 &\textbf{0.993} ± 0.015 &5.57  & 529.999 ± 201.308 \\
    \bottomrule
  \end{tabular}}

  \vspace{0.2em}

  \resizebox{\textwidth}{!}{
  \begin{tabular}{lcccccc}
    & \multicolumn{6}{c}{Dataset: \textbf{Analyst} \hspace{3mm} Task: \textbf{Event Anomaly}} \\
    \multirow{2}{*}{Method} & \multicolumn{4}{c}{Detection Effectiveness} & \multicolumn{2}{c}{Efficiency} \\
    \cmidrule(lr){2-5} \cmidrule(lr){6-7}
               & Acc. & Prec. & Recall & F1 & GFLOPs & Throughput (it/s) \\
    \midrule
    MLP &0.948 ± 0.019 &0.922 ± 0.064 &0.980 ± 0.043 &0.950 ± 0.016 &5.58 &1996.0 ± 454.0 \\
    RAG (Llama3-70b,$k$=5) &0.408 ± 0.038 &0.426 ± 0.026 &0.527 ± 0.029 &0.471 ± 0.019 &$\sim10^5$ &0.204 ± 0.009 \\
    RAG (GPT-oss-20b,$k$=5) &0.770 ± 0.106 &0.771 ± 0.157 &0.773 ± 0.014 &0.772 ± 0.085 &$\sim10^4$ &0.145 ± 0.018 \\
    RAG (Llama3-70b,$k$=10) &0.377 ± 0.014 &0.400 ± 0.004 &0.493 ± 0.038 &0.442 ± 0.014 &$\sim10^5$ &0.204 ± 0.015 \\
    RAG (GPT-oss-20b,$k$=10) &0.803 ± 0.090 &0.798 ± 0.099 &0.813 ± 0.087 &0.805 ± 0.087 &$\sim10^4$ &0.149 ± 0.014 \\
    \midrule
    GraphLogDebugger (Ours) &\textbf{0.957} ± 0.040 &\textbf{0.921} ± 0.069 &\textbf{1.000} ± 0.000 &\textbf{0.959} ± 0.037 &8.97 &1037.6 ± 286.2 \\
    \bottomrule
  \end{tabular}}

  \vspace{0.2em}

  \resizebox{\textwidth}{!}{
  \begin{tabular}{lcccccc}
    & \multicolumn{6}{c}{Dataset: \textbf{Landslide} \hspace{3mm} Task: \textbf{Object Anomaly}} \\
    \multirow{2}{*}{Method} & \multicolumn{4}{c}{Detection Effectiveness} & \multicolumn{2}{c}{Efficiency} \\
    \cmidrule(lr){2-5} \cmidrule(lr){6-7}
               & Acc. & Prec. & Recall & F1 & GFLOPs & Throughput (it/s) \\
    \midrule
    MLP &0.831 ± 0.079 &0.842 ± 0.180 &0.841 ± 0.149 &0.838 ± 0.056 & 5.58  &5391.0 ± 328.0 \\
    RAG (Llama3-70b,$k$=5) &0.543 ± 0.074 &\textbf{0.944} ± 0.056 &0.095 ± 0.156 &0.168 ± 0.267   &$\sim10^5$ &0.355 ± 0.044  \\
    RAG (GPT-oss-20b,$k$=5) &0.611 ± 0.068 &0.701 ± 0.050 &0.389 ± 0.246 &0.495 ± 0.195 &$\sim10^4$ & 0.155 ± 0.136 \\
    RAG (Llama3-70b,$k$=10) &0.551 ± 0.034 &0.904 ± 0.204 &0.119 ± 0.102 &0.208 ± 0.160 &$\sim10^5$ &0.321 ± 0.193 \\
    RAG (GPT-oss-20b,$k$=10) &0.623 ± 0.074 &0.723 ± 0.081 &0.397 ± 0.149 &0.511 ± 0.144 &$\sim10^4$ &0.166 ± 0.008 \\
    \midrule
    GraphLogDebugger (Ours) & \textbf{0.840} ± 0.080 &0.798 ± 0.117 &\textbf{0.929 } ± 0.059 &\textbf{0.858} ± 0.062 &19.70 &1334.13±108.40 \\
    \bottomrule
  \end{tabular}}



  \vspace{-0.4cm}
\end{table*}

\textbf{Baselines.} Our framework naturally generalizes to tabular log in different domains. Hence, we mainly compare it to baselines which are generally capable of dynamically processing different types of tabular log that contains text-rich and categorical attributes. \textbf{MLP} exploits a pretrained text embedding model to embed events and a learnable embedding for objects. A 3-layer MLP is then applied to map the embeddings to anomaly scores. We also compare a series of baselines based on retrieval augmented generation (\textbf{RAG})~\citep{lewis2020retrieval}, which is the mainstream method to process general tabular log in the realistic scenario~\citep{akhtar2025llm}. We deploy RAG based on two advanced open-sourced LLMs, Llama-3-70b~\citep{dubey2024llama} and GPT-oss-20b~\citep{agarwal2025gpt} with the 5 and 10 retrieval entries. The retrieval database is built on the whole training split and the seen log entries during the online evaluation. We construct specified prompts for different datasets and omitted the description to Appendix~\ref{appendix:A1}. Both MLP and all RAG baselines use all-MiniLM-L6-v2~\footnote{\url{https://huggingface.co/sentence-transformers/all-MiniLM-L6-v2}}~\citep{reimers-2019-sentence-bert} as the text embedding model.

\textbf{Task.} Following the setting of unsupervised anomaly detection~\citep{liu2021anomaly,schmidl2022anomaly}, our basic task is to output
an anomaly score for each log entry $x_{n}$ at timestamp $t_n$
, where higher scores denote more outlyingness~\citep{han2022adbench}. In our task, we use 1 to denote anomalies and 0 to denote normal examples in the ground-truth. We use the first 90\% split of the dataset for training, where both the log entries and their anomaly labels are available to access for methods. Methods train the model on this training split or use it for the retrieval database. For the rest 10\%, we use it as the test split in our online evaluation, where methods can make use of the seen log entries but their anomaly labels are not accessible. We study two types of anomalies in our experiments: object anomalies and event anomalies. Following the definition in Section~\ref{sec:3.1}, we inject object anomalies by swapping an object in the log entry with another existing object. To ensure that historical data contains useful information, we only perturb existing objects in the history. Event anomalies are generated similarly by swapping events. The anomaly rate is set to be 0.05. 

\textbf{Evaluation.} We calculate the metrics for information retrieval: accuracy, precision, recall, and f1 score for the dataset. We also evaluate the efficiency of different methods by GFlops and throughputs. During evaluation, we notice that LLM-based baselines tend to be very slow in processing speed. Hence, we include all anomaly log entries and 50 random normal entries in a subset and run RAG only on this subset. For other baselines and our methods, we obtain the prediction result for the full test split but only compute the metric on the above subset for fair comparison. We run experiments three times and post the average value of metrics with error bars.

\textbf{GraphLogDebugger.} The GNN architecture in GraphLogDebugger is a 3-layer GAT backbone with a two-branch node embedder and an MLP prediction head. The node embedder uses 512-d embeddings for objects and the text embedding of all-MiniLM-L6-v2~\citep{reimers-2019-sentence-bert} for events, with an MLP to map them into the same space. The embedding size of GAT and the prediction head is also 512. We train the GNN for 10 epochs under the learning rate 0.0001 on Adam and the negative ratio 10 on the training split. Following~\citet{you2022roland}, we set the batch size as 1 and use a window length of 100 to accelerate the processing.

\subsection{Main Results}
\label{sec:5.2}

Table~\ref{tab:res-detection-efficiency} shows that \textbf{GraphLogDebugger} consistently outperforms both MLP and RAG-based baselines across all tabular log datasets on five tasks, in terms of detection performance and efficiency.

\textbf{Effectiveness:} GraphLogDebugger achieves the highest F1 scores across all tasks, outperforming RAG baselines—especially in structurally complex domains like HDFS, where RAG methods fail to detect meaningful anomalies (F1 = 0.0). Specifically, RAG baselines are completely fooled by the anomaly pattern that their predicted labels depend on whether there is an existing record with the same format in the retrieved examples, which does not contribute to a reasonable prediction. Two tasks on the Arxiv dataset are the most difficult, where GraphLogDebugger still beats baselines with a higher precision in not abusing anomaly prediction. Even in semantically rich settings such as Analyst and LandSlide, where RAG baselines are expected to excel, our model surpasses them. 

\textbf{Efficiency:} RAG approaches exhibit extremely low throughput (typically below 0.3 iterations per second) due to the computational overhead of large language models. In contrast, GraphLogDebugger achieves throughput of at least 500 per second, with significantly lower GFLOPs, enabling real-time anomaly detection in high-throughput environments.

\subsection{Case study: Where does RAG fail?}

\begin{wrapfigure}{r}{0.4\linewidth}
\vspace{-0.3cm}
  \begin{tabular}{lc}
    \toprule
    Method & Correlation \\
    \midrule
    RAG & -0.1087 \\
    GraphLogDebugger & 0.1561 \\
    \bottomrule
  \end{tabular}
  \vspace{-0.2cm}
\end{wrapfigure}

It is natural that GraphLogDebugger yields advantages in efficiency compared to the RAG-baseline, for the latter relies on LLMs with billions of parameters. However, the leading performance of GraphLogDebugger in detection needs further explanation, while RAG enjoys the general comprehension and reasoning ability of modern LLMs. To this end, we study cases from event anomaly detection of the Arxiv dataset. We choose this task because the degree of event nodes can directly reflect the local graph density of the node-of-interest. We calculate the correlation between event node degrees and the accuracy of GraphLogDebugger and that of RAG(GPT-oss-20b,$k$=10). The result in the table shows that \textbf{the accuracy of GraphLogDebugger is positively correlated with the node degree, while the accuracy of RAG is negatively correlated with the node degree.} This indicates that GraphLogDebugger outperforms RAG on event nodes with rich connections with objects, where semantics of these objects are necessary to detect the anomaly.

\begin{casestudyframe}
\textbf{Case 1: Label=\textcolor{green}{negative}, RAG=\textcolor{red}{positive}, GraphLogDebugger=\textcolor{green}{negative}}\\
    \textbf{Title}: "SymbioSim: Human-in-the-loop Simulation Platform for Bidirectional Continuing Learning in Human-Robot Interaction"\\
    \textbf{Authors}: "Haoran Chen", "Yiming Ren", "Xinran Li", "Ning Ding", "Ziyi Wang", "Yuhan Chen", "Zhiyang Dou", "Yuexin Ma", "Changhe Tu" (9 objects)\\
    \textbf{Reason (RAG)}: The author team composition, research domain mismatch, and unclear collaboration patterns raise suspicions about the coherence of the record.\\
\textbf{Case 2: Label=\textcolor{red}{positive}, RAG=\textcolor{green}{negative}, GraphLogDebugger=\textcolor{red}{positive}}\\
\textbf{Title}: "VERA: Explainable Video Anomaly Detection via Verbalized Learning of Vision-Language Models"\\
\textbf{Authors}: "Shubham Gupta", "Zichao Li", "Tianyi Chen", "Cem Subakan", "Siva Reddy", "Perouz Taslakian", "Valentina Zantedeschi" (7 objects)\\
\textbf{Reason (RAG)}: The record seems coherent, with individual authors' expertise areas aligning with the paper's topic, although the team size is slightly larger than expected.\\
\textbf{Case 3: Label=\textcolor{red}{positive}, RAG=\textcolor{red}{positive}, GraphLogDebugger=\textcolor{green}{negative}}\\
\textbf{Title}: "Transformer$^{-1}$: Input-Adaptive Computation for Resource-Constrained Deployment"\\
\textbf{Authors}: "Yitong Yin" (1 objects)\\
\textbf{Reason (RAG)}: The record consists of a single author, which is consistent with similar papers in the same research domain.
\end{casestudyframe}
We further raise three cases above to investigate when and how GraphLogDebugger and RAG fail. In Case 1, RAG predicts the normal example as abnormal because the limited retrieved examples do not provide enough evidence to prove the coherence of the author team. By contrast, GraphLogDebugger validates overall team consistency by checking the research background of every author, which correctly predicts the negative label. Case 2 is complementary to Case 1, where GraphLogDebugger is able to scan the research interest of every author and detect the anomaly accurately. However, when the connected objects are few, such as in Case 3, GraphLogDebugger may not have enough references based on the graph to make a correct judgment. In similar cases, RAG could then outperform GraphLogDebugger to recognize patterns in the number of authors in the same domain.

These cases provide insights on how graphs can benefit retrieval augmented generation. When the key entry has dense connections with other entries, traditional retrieval based on similarity cannot efficiently include enough entries to enhance the generation quality. With the help of modern embedding models, graphs can be introduced to gather information in these multi-entry scenarios.




\section{Conclusion}
We propose a general framework to cover online debugging for heterogeneous tabular logs. By modeling online log debugging as anomaly detection of dynamic graphs, our framework integrates different types of log data into a unified modality by text embedding models, where a dynamic GNN debugs the log through link prediction. Our framework shows good performance in four different datasets while maintaining high efficiency compared to the mainstream RAG-based method.

\textbf{Limitation.} Our work explores combining dynamic GNNs and text embedding models to process log data under the online setting, which indicates the potential to accelerate the online process of data streams by a graph-based method. Nevertheless, our experiments mainly show this potential in the bug detection setting. We leave the exploration of online bug correction to future work.

\section*{Acknowledgments}
We sincerely appreciate the support from Amazon grant funding project \#120359, "GRAG: Enhance RAG Applications with Graph-structured Knowledge".

\section*{Ethics statement}
Our work focuses on detecting inconsistencies in general tabular log data, which enhances the progress of automated log data processing in real-world scenarios. While automation of log processing may raise issues concerning hallucination or fraud reporting, our work does not explicitly introduce new risks compared to existing research.
\section*{Reproducibility statement}
All implementation details of our method and baselines are given in Section~\ref{sec:5} and Appendix~\ref{appendix:A1}. We will release at the time of publication. 

\bibliography{iclr2026_conference}

@article{liu2021anomaly,
  title={Anomaly detection on attributed networks via contrastive self-supervised learning},
  author={Liu, Yixin and Li, Zhao and Pan, Shirui and Gong, Chen and Zhou, Chuan and Karypis, George},
  journal={IEEE transactions on neural networks and learning systems},
  volume={33},
  number={6},
  pages={2378--2392},
  year={2021},
  publisher={IEEE}
}

@inproceedings{koval2024financial,
  title={Financial forecasting from textual and tabular time series},
  author={Koval, Ross and Andrews, Nicholas and Yan, Xifeng},
  booktitle={Findings of the Association for Computational Linguistics: EMNLP 2024},
  pages={8289--8300},
  year={2024}
}

@inproceedings{ruiz2012correlating,
  title={Correlating financial time series with micro-blogging activity},
  author={Ruiz, Eduardo J and Hristidis, Vagelis and Castillo, Carlos and Gionis, Aristides and Jaimes, Alejandro},
  booktitle={Proceedings of the fifth ACM international conference on Web search and data mining},
  pages={513--522},
  year={2012}
}

@article{tetlock2007giving,
  title={Giving content to investor sentiment: The role of media in the stock market},
  author={Tetlock, Paul C},
  journal={The Journal of finance},
  volume={62},
  number={3},
  pages={1139--1168},
  year={2007},
  publisher={Wiley Online Library}
}

@misc{dong2024fnspid,
  title={FNSPID: A Comprehensive Financial News Dataset in Time Series},
  author={Zihan Dong and Xinyu Fan and Zhiyuan Peng},
  year={2024},
  eprint={2402.06698},
  archivePrefix={arXiv},
  primaryClass={q-fin.ST}
}

@article{clement2019use,
  title={On the use of arxiv as a dataset},
  author={Clement, Colin B and Bierbaum, Matthew and O'Keeffe, Kevin P and Alemi, Alexander A},
  journal={arXiv preprint arXiv:1905.00075},
  year={2019}
}

@article{kinney2023semantic,
  title={The semantic scholar open data platform},
  author={Kinney, Rodney and Anastasiades, Chloe and Authur, Russell and Beltagy, Iz and Bragg, Jonathan and Buraczynski, Alexandra and Cachola, Isabel and Candra, Stefan and Chandrasekhar, Yoganand and Cohan, Arman and others},
  journal={arXiv preprint arXiv:2301.10140},
  year={2023}
}

@article{johnson2023mimic,
  title={MIMIC-IV, a freely accessible electronic health record dataset},
  author={Johnson, Alistair EW and Bulgarelli, Lucas and Shen, Lu and Gayles, Alvin and Shammout, Ayad and Horng, Steven and Pollard, Tom J and Hao, Sicheng and Moody, Benjamin and Gow, Brian and others},
  journal={Scientific data},
  volume={10},
  number={1},
  pages={1},
  year={2023},
  publisher={Nature Publishing Group UK London}
}

@inproceedings{zhu2023loghub,
  title={Loghub: A large collection of system log datasets for ai-driven log analytics},
  author={Zhu, Jieming and He, Shilin and He, Pinjia and Liu, Jinyang and Lyu, Michael R},
  booktitle={2023 IEEE 34th International Symposium on Software Reliability Engineering (ISSRE)},
  pages={355--366},
  year={2023},
  organization={IEEE}
}

@article{wu2025tabular,
  title={Tabular Data Understanding with LLMs: A Survey of Recent Advances and Challenges},
  author={Wu, Xiaofeng and Ritter, Alan and Xu, Wei},
  journal={arXiv preprint arXiv:2508.00217},
  year={2025}
}

@article{kurisinkel2024text2timeseries,
  title={Text2timeseries: Enhancing financial forecasting through time series prediction updates with event-driven insights from large language models},
  author={Kurisinkel, Litton Jose and Mishra, Pruthwik and Zhang, Yue},
  journal={arXiv preprint arXiv:2407.03689},
  year={2024}
}

@article{xiao2025enhancing,
  title={Enhancing Financial Time-Series Forecasting with Retrieval-Augmented Large Language Models},
  author={Xiao, Mengxi and Jiang, Zihao and Qian, Lingfei and Chen, Zhengyu and He, Yueru and Xu, Yijing and Jiang, Yuecheng and Li, Dong and Weng, Ruey-Ling and Peng, Min and others},
  journal={arXiv preprint arXiv:2503.67890},
  year={2025}
}

@article{zhao2022utilizing,
  title={Utilizing citation network structure to predict paper citation counts: A deep learning approach},
  author={Zhao, Qihang and Feng, Xiaodong},
  journal={Journal of Informetrics},
  volume={16},
  number={1},
  pages={101235},
  year={2022},
  publisher={Elsevier}
}

@inproceedings{pang2025guard,
  title={GuARD: Effective Anomaly Detection through a Text-Rich and Graph-Informed Language Model},
  author={Pang, Yunhe and Chen, Bo and Zhang, Fanjin and Rao, Yanghui and Kharlamov, Evgeny and Tang, Jie},
  booktitle={Proceedings of the 31st ACM SIGKDD Conference on Knowledge Discovery and Data Mining V. 2},
  pages={2222--2233},
  year={2025}
}

@article{yu2023temporal,
  title={Temporal data meets LLM--explainable financial time series forecasting},
  author={Yu, Xinli and Chen, Zheng and Ling, Yuan and Dong, Shujing and Liu, Zongyi and Lu, Yanbin},
  journal={arXiv preprint arXiv:2306.11025},
  year={2023}
}

@article{fang2024large,
  title={Large Language Models (LLMs) on Tabular Data: Prediction, Generation, and Understanding--A Survey},
  author={Fang, Xi and Xu, Weijie and Tan, Fiona Anting and Zhang, Jiani and Hu, Ziqing and Qi, Yanjun and Nickleach, Scott and Socolinsky, Diego and Sengamedu, Srinivasan and Faloutsos, Christos},
  journal={arXiv preprint arXiv:2402.17944},
  year={2024}
}

@article{wang2024chain,
  title={Chain-of-table: Evolving tables in the reasoning chain for table understanding},
  author={Wang, Zilong and Zhang, Hao and Li, Chun-Liang and Eisenschlos, Julian Martin and Perot, Vincent and Wang, Zifeng and Miculicich, Lesly and Fujii, Yasuhisa and Shang, Jingbo and Lee, Chen-Yu and others},
  journal={arXiv preprint arXiv:2401.04398},
  year={2024}
}

@article{brown2020language,
  title={Language models are few-shot learners},
  author={Brown, Tom and Mann, Benjamin and Ryder, Nick and Subbiah, Melanie and Kaplan, Jared D and Dhariwal, Prafulla and Neelakantan, Arvind and Shyam, Pranav and Sastry, Girish and Askell, Amanda and others},
  journal={Advances in neural information processing systems},
  volume={33},
  pages={1877--1901},
  year={2020}
}

@article{liu2023rethinking,
  title={Rethinking tabular data understanding with large language models},
  author={Liu, Tianyang and Wang, Fei and Chen, Muhao},
  journal={arXiv preprint arXiv:2312.16702},
  year={2023}
}

@article{zhang2024tablellm,
  title={Tablellm: Enabling tabular data manipulation by llms in real office usage scenarios},
  author={Zhang, Xiaokang and Luo, Sijia and Zhang, Bohan and Ma, Zeyao and Zhang, Jing and Li, Yang and Li, Guanlin and Yao, Zijun and Xu, Kangli and Zhou, Jinchang and others},
  journal={arXiv preprint arXiv:2403.19318},
  year={2024}
}

@inproceedings{zhong2024logparser,
  title={Logparser-llm: Advancing efficient log parsing with large language models},
  author={Zhong, Aoxiao and Mo, Dengyao and Liu, Guiyang and Liu, Jinbu and Lu, Qingda and Zhou, Qi and Wu, Jiesheng and Li, Quanzheng and Wen, Qingsong},
  booktitle={Proceedings of the 30th ACM SIGKDD Conference on Knowledge Discovery and Data Mining},
  pages={4559--4570},
  year={2024}
}

@article{gardner2024large,
  title={Large scale transfer learning for tabular data via language modeling},
  author={Gardner, Josh and Perdomo, Juan C and Schmidt, Ludwig},
  journal={Advances in Neural Information Processing Systems},
  volume={37},
  pages={45155--45205},
  year={2024}
}

@article{wang2024news,
  title={From news to forecast: Integrating event analysis in llm-based time series forecasting with reflection},
  author={Wang, Xinlei and Feng, Maike and Qiu, Jing and Gu, Jinjin and Zhao, Junhua},
  journal={Advances in Neural Information Processing Systems},
  volume={37},
  pages={58118--58153},
  year={2024}
}

@inproceedings{yang2018dcfee,
  title={Dcfee: A document-level chinese financial event extraction system based on automatically labeled training data},
  author={Yang, Hang and Chen, Yubo and Liu, Kang and Xiao, Yang and Zhao, Jun},
  booktitle={Proceedings of ACL 2018, System Demonstrations},
  pages={50--55},
  year={2018}
}

@article{ye2024mirai,
  title={Mirai: Evaluating llm agents for event forecasting},
  author={Ye, Chenchen and Hu, Ziniu and Deng, Yihe and Huang, Zijie and Ma, Mingyu Derek and Zhu, Yanqiao and Wang, Wei},
  journal={arXiv preprint arXiv:2407.01231},
  year={2024}
}

@article{shi2023language,
  title={Language models can improve event prediction by few-shot abductive reasoning},
  author={Shi, Xiaoming and Xue, Siqiao and Wang, Kangrui and Zhou, Fan and Zhang, James and Zhou, Jun and Tan, Chenhao and Mei, Hongyuan},
  journal={Advances in Neural Information Processing Systems},
  volume={36},
  pages={29532--29557},
  year={2023}
}

@inproceedings{li2023glad,
  title={Glad: Content-aware dynamic graphs for log anomaly detection},
  author={Li, Yufei and Liu, Yanchi and Wang, Haoyu and Chen, Zhengzhang and Cheng, Wei and Chen, Yuncong and Yu, Wenchao and Chen, Haifeng and Liu, Cong},
  booktitle={2023 IEEE International Conference on Knowledge Graph (ICKG)},
  pages={9--18},
  year={2023},
  organization={IEEE}
}

@article{kipf2016semi,
  title={Semi-supervised classification with graph convolutional networks},
  author={Kipf, TN},
  journal={arXiv preprint arXiv:1609.02907},
  year={2016}
}

@article{hamilton2017inductive,
  title={Inductive representation learning on large graphs},
  author={Hamilton, Will and Ying, Zhitao and Leskovec, Jure},
  journal={Advances in neural information processing systems},
  volume={30},
  year={2017}
}

@inproceedings{gilmer2017neural,
  title={Neural message passing for quantum chemistry},
  author={Gilmer, Justin and Schoenholz, Samuel S and Riley, Patrick F and Vinyals, Oriol and Dahl, George E},
  booktitle={International conference on machine learning},
  pages={1263--1272},
  year={2017},
  organization={Pmlr}
}

@article{battaglia2018relational,
  title={Relational inductive biases, deep learning, and graph networks},
  author={Battaglia, Peter W and Hamrick, Jessica B and Bapst, Victor and Sanchez-Gonzalez, Alvaro and Zambaldi, Vinicius and Malinowski, Mateusz and Tacchetti, Andrea and Raposo, David and Santoro, Adam and Faulkner, Ryan and others},
  journal={arXiv preprint arXiv:1806.01261},
  year={2018}
}

@article{li2021deepgcns,
  title={DeepGCNs: Can GCNs go as deep as CNNs?},
  author={Li, Guohao and Müller, Matthias and Thabet, Ali and Ghanem, Bernard},
  journal={IEEE Transactions on Pattern Analysis and Machine Intelligence},
  year={2021}
}

@article{dwivedi2020benchmarking,
  title={Benchmarking graph neural networks},
  author={Dwivedi, Vijay Prakash and Joshi, Chaitanya K and Laurent, Thomas and Bengio, Yoshua and Bresson, Xavier},
  journal={arXiv preprint arXiv:2003.00982},
  year={2020}
}

@article{li2017diffusion,
  title={Diffusion convolutional recurrent neural network: Data-driven traffic forecasting},
  author={Li, Yaguang and Yu, Rose and Shahabi, Cyrus and Liu, Yan},
  journal={arXiv preprint arXiv:1707.01926},
  year={2017}
}

@inproceedings{seo2018structured,
  title={Structured sequence modeling with graph convolutional recurrent networks},
  author={Seo, Youngjoo and Defferrard, Micha{\"e}l and Vandergheynst, Pierre and Bresson, Xavier},
  booktitle={International conference on neural information processing},
  pages={362--373},
  year={2018},
  organization={Springer}
}

@inproceedings{pareja2020evolvegcn,
  title={Evolvegcn: Evolving graph convolutional networks for dynamic graphs},
  author={Pareja, Aldo and Domeniconi, Giacomo and Chen, Jie and Ma, Tengfei and Suzumura, Toyotaro and Kanezashi, Hiroki and Kaler, Tim and Schardl, Tao and Leiserson, Charles},
  booktitle={Proceedings of the AAAI conference on artificial intelligence},
  volume={34},
  number={04},
  pages={5363--5370},
  year={2020}
}

@inproceedings{sankar2020dysat,
  title={Dysat: Deep neural representation learning on dynamic graphs via self-attention networks},
  author={Sankar, Aravind and Wu, Yanhong and Gou, Liang and Zhang, Wei and Yang, Hao},
  booktitle={Proceedings of the 13th international conference on web search and data mining},
  pages={519--527},
  year={2020}
}

@inproceedings{kumar2019predicting,
  title={Predicting dynamic embedding trajectory in temporal interaction networks},
  author={Kumar, Srijan and Zhang, Xikun and Leskovec, Jure},
  booktitle={Proceedings of the 25th ACM SIGKDD international conference on knowledge discovery \& data mining},
  pages={1269--1278},
  year={2019}
}

@article{xu2020inductive,
  title={Inductive representation learning on temporal graphs},
  author={Xu, Da and Ruan, Chuanwei and Korpeoglu, Evren and Kumar, Sushant and Achan, Kannan},
  journal={arXiv preprint arXiv:2002.07962},
  year={2020}
}

@article{rossi2020temporal,
  title={Temporal graph networks for deep learning on dynamic graphs},
  author={Rossi, Emanuele and Chamberlain, Ben and Frasca, Fabrizio and Eynard, Davide and Monti, Federico and Bronstein, Michael},
  journal={arXiv preprint arXiv:2006.10637},
  year={2020}
}

@inproceedings{you2022roland,
  title={ROLAND: graph learning framework for dynamic graphs},
  author={You, Jiaxuan and Du, Tianyu and Leskovec, Jure},
  booktitle={Proceedings of the 28th ACM SIGKDD conference on knowledge discovery and data mining},
  pages={2358--2366},
  year={2022}
}

@article{longa2023temporal,
  title={Graph neural networks for temporal graphs: State of the art, open challenges, and opportunities},
  author={Longa, Alessandro and Lachi, Valerio and Santin, Giovanni and Bianchini, Monica and Lepri, Bruno and others},
  journal={arXiv preprint arXiv:2305.12472},
  year={2023}
}

@article{huang2023tgb,
  title={Temporal graph benchmark for machine learning on temporal graphs},
  author={Huang, Shenyang and Poursafaei, Farimah and Danovitch, Jacob and Fey, Matthias and Hu, Weihua and Rossi, Emanuele and Leskovec, Jure and Bronstein, Michael and Rabusseau, Guillaume and Rabbany, Reihaneh},
  journal={Advances in Neural Information Processing Systems},
  volume={36},
  pages={2056--2073},
  year={2023}
}

@inproceedings{zhu2023wingnn,
  title={Wingnn: Dynamic graph neural networks with random gradient aggregation window},
  author={Zhu, Yifan and Cong, Fangpeng and Zhang, Dan and Gong, Wenwen and Lin, Qika and Feng, Wenzheng and Dong, Yuxiao and Tang, Jie},
  booktitle={Proceedings of the 29th ACM SIGKDD conference on knowledge discovery and data mining},
  pages={3650--3662},
  year={2023}
}

@article{chen2023tempme,
  title={Tempme: Towards the explainability of temporal graph neural networks via motif discovery},
  author={Chen, Jialin and Ying, Rex},
  journal={Advances in Neural Information Processing Systems},
  volume={36},
  pages={29005--29028},
  year={2023}
}

@inproceedings{song2023fair,
  title={Towards fair financial services for all: A temporal GNN approach for individual fairness on transaction networks},
  author={Song, Zixing and Zhang, Yuji and King, Irwin},
  booktitle={Proceedings of the 32nd ACM international conference on information and knowledge management},
  pages={2331--2341},
  year={2023}
}

@article{zhang2023spectral,
  title={Spectral invariant learning for dynamic graphs under distribution shifts},
  author={Zhang, Zeyang and Wang, Xin and Zhang, Ziwei and Qin, Zhou and Wen, Weigao and Xue, Hui and Li, Haoyang and Zhu, Wenwu},
  journal={Advances in Neural Information Processing Systems},
  volume={36},
  pages={6619--6633},
  year={2023}
}

@article{zhang2024dtgb,
  title={DTGB: A comprehensive benchmark for dynamic text-attributed graphs},
  author={Zhang, Jiasheng and Chen, Jialin and Yang, Menglin and Feng, Aosong and Liang, Shuang and Shao, Jie and Ying, Rex},
  journal={Advances in Neural Information Processing Systems},
  volume={37},
  pages={91405--91429},
  year={2024}
}

@inproceedings{xu2009detecting,
  title={Detecting large-scale system problems by mining console logs},
  author={Xu, Wei and Huang, Ling and Fox, Armando and Patterson, David and Jordan, Michael I},
  booktitle={Proceedings of the ACM SIGOPS 22nd symposium on Operating systems principles},
  pages={117--132},
  year={2009}
}

@inproceedings{du2017deeplog,
  title={Deeplog: Anomaly detection and diagnosis from system logs through deep learning},
  author={Du, Min and Li, Feifei and Zheng, Guineng and Srikumar, Vivek},
  booktitle={Proceedings of the 2017 ACM SIGSAC conference on computer and communications security},
  pages={1285--1298},
  year={2017}
}

@inproceedings{lu2018cnn,
  title={Detecting anomaly in big data system logs using convolutional neural network},
  author={Lu, Siyang and Wei, Xiang and Li, Yandong and Wang, Liqiang},
  booktitle={2018 IEEE 16th Intl Conf on Dependable, Autonomic and Secure Computing, 16th Intl Conf on Pervasive Intelligence and Computing, 4th Intl Conf on Big Data Intelligence and Computing and Cyber Science and Technology Congress (DASC/PiCom/DataCom/CyberSciTech)},
  pages={151--158},
  year={2018},
  organization={IEEE}
}

@inproceedings{zhang2021logattn,
  title={LogAttn: Unsupervised log anomaly detection with an autoencoder based attention mechanism},
  author={Zhang, Linming and Li, Wenzhong and Zhang, Zhijie and Lu, Qingning and Hou, Ce and Hu, Peng and Gui, Tong and Lu, Sanglu},
  booktitle={International conference on knowledge science, engineering and management},
  pages={222--235},
  year={2021},
  organization={Springer}
}

@article{castillo2022autolog,
  title={Autolog: Anomaly detection by deep autoencoding of system logs},
  author={Castillo, Martina and Pecchia, Antonio and Villano, Ugo},
  journal={Expert Systems with Applications},
  volume={191},
  year={2022}
}

@inproceedings{zhang2023vae,
  title={Anomaly detection model for log based on lstm network and variational autoencoder},
  author={Zhang, Xinye and Chai, Xiaoli and Yu, Minghua and Qiu, Ding},
  booktitle={2023 4th International Conference on Information Science, Parallel and Distributed Systems (ISPDS)},
  pages={239--244},
  year={2023},
  organization={IEEE}
}

@article{duan2021gan,
  title={A Generative Adversarial Networks for Log Anomaly Detection.},
  author={Duan, Xiaoyu and Ying, Shi and Yuan, Wanli and Cheng, Hailong and Yin, Xiang},
  journal={Computer Systems Science \& Engineering},
  volume={37},
  number={1},
  year={2021}
}

@inproceedings{he2023gan,
  title={Graph-based log anomaly detection via adversarial training},
  author={He, Zhangyue and Tang, Yanni and Zhao, Kaiqi and Liu, Jiamou and Chen, Wu},
  booktitle={International Symposium on Dependable Software Engineering: Theories, Tools, and Applications},
  pages={55--71},
  year={2023},
  organization={Springer}
}

@inproceedings{zhang2019logrobust,
  title={Robust log-based anomaly detection on unstable log data},
  author={Zhang, Xu and Xu, Yong and Lin, Qingwei and Qiao, Bo and Zhang, Hongyu and Dang, Yingnong and Xie, Chunyu and Yang, Xinsheng and Cheng, Qian and Li, Ze and others},
  booktitle={Proceedings of the 2019 27th ACM joint meeting on European software engineering conference and symposium on the foundations of software engineering},
  pages={807--817},
  year={2019}
}

@inproceedings{yang2021plelog,
  title={Semi-supervised log-based anomaly detection via probabilistic label estimation},
  author={Yang, Lin and Chen, Junjie and Wang, Zan and Wang, Weijing and Jiang, Jiajun and Dong, Xuyuan and Zhang, Wenbin},
  booktitle={2021 IEEE/ACM 43rd International Conference on Software Engineering (ICSE)},
  pages={1448--1460},
  year={2021},
  organization={IEEE}
}

@inproceedings{guo2021logbert,
  title={Logbert: Log anomaly detection via bert},
  author={Guo, Haixuan and Yuan, Shuhan and Wu, Xintao},
  booktitle={2021 international joint conference on neural networks (IJCNN)},
  pages={1--8},
  year={2021},
  organization={IEEE}
}

@article{lee2023lanobert,
  title={Lanobert: System log anomaly detection based on bert masked language model},
  author={Lee, Yukyung and Kim, Jina and Kang, Pilsung},
  journal={Applied Soft Computing},
  volume={146},
  pages={110689},
  year={2023},
  publisher={Elsevier}
}

@inproceedings{qi2023loggpt,
  title={Loggpt: Exploring chatgpt for log-based anomaly detection},
  author={Qi, Jiaxing and Huang, Shaohan and Luan, Zhongzhi and Yang, Shu and Fung, Carol and Yang, Hailong and Qian, Depei and Shang, Jing and Xiao, Zhiwen and Wu, Zhihui},
  booktitle={2023 IEEE International Conference on High Performance Computing \& Communications, Data Science \& Systems, Smart City \& Dependability in Sensor, Cloud \& Big Data Systems \& Application (HPCC/DSS/SmartCity/DependSys)},
  pages={273--280},
  year={2023},
  organization={IEEE}
}

@inproceedings{liu2024logprompt,
  title={Logprompt: Prompt engineering towards zero-shot and interpretable log analysis},
  author={Liu, Yilun and Tao, Shimin and Meng, Weibin and Yao, Feiyu and Zhao, Xiaofeng and Yang, Hao},
  booktitle={Proceedings of the 2024 IEEE/ACM 46th International Conference on Software Engineering: Companion Proceedings},
  pages={364--365},
  year={2024}
}

@inproceedings{pan2024raglog,
  title={Raglog: Log anomaly detection using retrieval augmented generation},
  author={Pan, Jonathan and Liang, Wong Swee and Yidi, Yuan},
  booktitle={2024 IEEE World Forum on Public Safety Technology (WFPST)},
  pages={169--174},
  year={2024},
  organization={IEEE}
}

@article{no2024rapid,
  title={Training-free retrieval-based log anomaly detection with pre-trained language models},
  author={No, Geon and others},
  journal={Engineering Applications of Artificial Intelligence},
  volume={133},
  year={2024}
}

@inproceedings{meng2019loganomaly,
  title={Loganomaly: Unsupervised detection of sequential and quantitative anomalies in unstructured logs.},
  author={Meng, Weibin and Liu, Ying and Zhu, Yichen and Zhang, Shenglin and Pei, Dan and Liu, Yuqing and Chen, Yihao and Zhang, Ruizhi and Tao, Shimin and Sun, Pei and others},
  booktitle={IJCAI},
  volume={19},
  number={7},
  pages={4739--4745},
  year={2019}
}

@article{feng2025graph,
  title={Graph World Model},
  author={Feng, Tao and Wu, Yexin and Lin, Guanyu and You, Jiaxuan},
  journal={arXiv preprint arXiv:2507.10539},
  year={2025}
}

@article{zhang2025agent,
  title={Which agent causes task failures and when? on automated failure attribution of llm multi-agent systems},
  author={Zhang, Shaokun and Yin, Ming and Zhang, Jieyu and Liu, Jiale and Han, Zhiguang and Zhang, Jingyang and Li, Beibin and Wang, Chi and Wang, Huazheng and Chen, Yiran and others},
  journal={arXiv preprint arXiv:2505.00212},
  year={2025}
}

@article{ekle2024anomaly,
  title={Anomaly detection in dynamic graphs: A comprehensive survey},
  author={Ekle, Ocheme Anthony and Eberle, William},
  journal={ACM Transactions on Knowledge Discovery from Data},
  volume={18},
  number={8},
  pages={1--44},
  year={2024},
  publisher={ACM New York, NY}
}

@article{pang2021deep,
  title={Deep learning for anomaly detection: A review},
  author={Pang, Guansong and Shen, Chunhua and Cao, Longbing and Hengel, Anton Van Den},
  journal={ACM computing surveys (CSUR)},
  volume={54},
  number={2},
  pages={1--38},
  year={2021},
  publisher={ACM New York, NY, USA}
}

@article{velivckovic2017graph,
  title={Graph attention networks},
  author={Veli{\v{c}}kovi{\'c}, Petar and Cucurull, Guillem and Casanova, Arantxa and Romero, Adriana and Lio, Pietro and Bengio, Yoshua},
  journal={arXiv preprint arXiv:1710.10903},
  year={2017}
}

@article{schmidl2022anomaly,
  title={Anomaly detection in time series: a comprehensive evaluation},
  author={Schmidl, Sebastian and Wenig, Phillip and Papenbrock, Thorsten},
  journal={Proceedings of the VLDB Endowment},
  volume={15},
  number={9},
  pages={1779--1797},
  year={2022},
  publisher={VLDB Endowment}
}

@article{han2022adbench,
  title={Adbench: Anomaly detection benchmark},
  author={Han, Songqiao and Hu, Xiyang and Huang, Hailiang and Jiang, Minqi and Zhao, Yue},
  journal={Advances in neural information processing systems},
  volume={35},
  pages={32142--32159},
  year={2022}
}

@article{lewis2020retrieval,
  title={Retrieval-augmented generation for knowledge-intensive nlp tasks},
  author={Lewis, Patrick and Perez, Ethan and Piktus, Aleksandra and Petroni, Fabio and Karpukhin, Vladimir and Goyal, Naman and K{\"u}ttler, Heinrich and Lewis, Mike and Yih, Wen-tau and Rockt{\"a}schel, Tim and others},
  journal={Advances in neural information processing systems},
  volume={33},
  pages={9459--9474},
  year={2020}
}

@article{akhtar2025llm,
  title={LLM-based event log analysis techniques: A survey},
  author={Akhtar, Siraaj and Khan, Saad and Parkinson, Simon},
  journal={arXiv preprint arXiv:2502.00677},
  year={2025}
}

@article{agarwal2025gpt,
  title={gpt-oss-120b \& gpt-oss-20b model card},
  author={Agarwal, Sandhini and Ahmad, Lama and Ai, Jason and Altman, Sam and Applebaum, Andy and Arbus, Edwin and Arora, Rahul K and Bai, Yu and Baker, Bowen and Bao, Haiming and others},
  journal={arXiv preprint arXiv:2508.10925},
  year={2025}
}

@article{dubey2024llama,
  title={The llama 3 herd of models},
  author={Dubey, Abhimanyu and Jauhri, Abhinav and Pandey, Abhinav and Kadian, Abhishek and Al-Dahle, Ahmad and Letman, Aiesha and Mathur, Akhil and Schelten, Alan and Yang, Amy and Fan, Angela and others},
  journal={arXiv e-prints},
  pages={arXiv--2407},
  year={2024}
}

@inproceedings{reimers-2019-sentence-bert,
    title = "Sentence-BERT: Sentence Embeddings using Siamese BERT-Networks",
    author = "Reimers, Nils and Gurevych, Iryna",
    booktitle = "Proceedings of the 2019 Conference on Empirical Methods in Natural Language Processing",
    month = "11",
    year = "2019",
    publisher = "Association for Computational Linguistics",
    url = "https://arxiv.org/abs/1908.10084",
}

@inproceedings{cheng2020knowledge,
  title={Knowledge graph-based event embedding framework for financial quantitative investments},
  author={Cheng, Dawei and Yang, Fangzhou and Wang, Xiaoyang and Zhang, Ying and Zhang, Liqing},
  booktitle={Proceedings of the 43rd International ACM SIGIR Conference on Research and Development in Information Retrieval},
  pages={2221--2230},
  year={2020}
}

@article{zehra2021financial,
  title={Financial knowledge graph based financial report query system},
  author={Zehra, Samreen and Mohsin, Syed Farhan Mohsin and Wasi, Shaukat and Jami, Syed Imran and Siddiqui, Muhammad Shoaib and Syed, Muhammad Khaliq-Ur-Rahman Raazi},
  journal={IEEE Access},
  volume={9},
  pages={69766--69782},
  year={2021},
  publisher={IEEE}
}

@article{sui2023logkg,
  title={Logkg: Log failure diagnosis through knowledge graph},
  author={Sui, Yicheng and Zhang, Yuzhe and Sun, Jianjun and Xu, Ting and Zhang, Shenglin and Li, Zhengdan and Sun, Yongqian and Guo, Fangrui and Shen, Junyu and Zhang, Yuzhi and others},
  journal={IEEE Transactions on Services Computing},
  volume={16},
  number={5},
  pages={3493--3507},
  year={2023},
  publisher={IEEE}
}

@article{hogan2021knowledge,
  title={Knowledge graphs},
  author={Hogan, Aidan and Blomqvist, Eva and Cochez, Michael and d’Amato, Claudia and Melo, Gerard De and Gutierrez, Claudio and Kirrane, Sabrina and Gayo, Jos{\'e} Emilio Labra and Navigli, Roberto and Neumaier, Sebastian and others},
  journal={ACM Computing Surveys (Csur)},
  volume={54},
  number={4},
  pages={1--37},
  year={2021},
  publisher={ACM New York, NY, USA}
}

@inproceedings{zhang2025xraglog,
  title={XRAGLog: A Resource-Efficient and Context-Aware Log-Based Anomaly Detection Method Using Retrieval-Augmented Generation},
  author={Zhang, Lingzhe and Jia, Tong and Jia, Mengxi and Wu, Yifan and Liu, Hongyi and Li, Ying},
  booktitle={AAAI 2025 Workshop on Preventing and Detecting LLM Misinformation (PDLM)},
  year={2025}
}

@inproceedings{pei2020subgraph,
  title={Subgraph anomaly detection in financial transaction networks},
  author={Pei, Yulong and Lyu, Fang and Van Ipenburg, Werner and Pechenizkiy, Mykola},
  booktitle={Proceedings of the First ACM International Conference on AI in Finance},
  pages={1--8},
  year={2020}
}

@inproceedings{chen2022antibenford,
  title={Antibenford subgraphs: Unsupervised anomaly detection in financial networks},
  author={Chen, Tianyi and Tsourakakis, Charalampos},
  booktitle={Proceedings of the 28th ACM SIGKDD Conference on Knowledge Discovery and Data Mining},
  pages={2762--2770},
  year={2022}
}

@article{park2024enhancing,
  title={Enhancing anomaly detection in financial markets with an llm-based multi-agent framework},
  author={Park, Taejin},
  journal={arXiv preprint arXiv:2403.19735},
  year={2024}
}

@inproceedings{he2016experience,
  title={Experience report: System log analysis for anomaly detection},
  author={He, Shilin and Zhu, Jieming and He, Pinjia and Lyu, Michael R},
  booktitle={2016 IEEE 27th international symposium on software reliability engineering (ISSRE)},
  pages={207--218},
  year={2016},
  organization={IEEE}
}

@article{wang2025financial,
  title={Financial analysis: Intelligent financial data analysis system based on llm-rag},
  author={Wang, Jingru and Ding, Wen and Zhu, Xiaotong},
  journal={arXiv preprint arXiv:2504.06279},
  year={2025}
}
\bibliographystyle{iclr2026_conference}

\appendix
\section{Appendix}
\label{appendix:A1}

\subsection{Use of LLMs}

We use ChatGPT to polish our introduction (Section~\ref{sec:1}) and generate the notation table (Table~\ref{tab:notation}), both of which have been checked manually. We also use ChatGPT to retrieve related works in the tabular log processing part by searching machine-learning based log processing methods.

\subsection{Notation}
\label{appendix:notation}
\begin{longtable}{@{}llp{0.54\textwidth}@{}}
\caption{\textbf{Notation}}
\label{tab:notation}\\
\toprule
\textbf{Symbol} & \textbf{Type} & \textbf{Meaning} \\
\midrule
\endfirsthead
\toprule
\textbf{Symbol} & \textbf{Type} & \textbf{Meaning} \\
\midrule
\endhead
\midrule
\multicolumn{3}{r}{\emph{Continued on next page}}\\
\bottomrule
\endfoot
\bottomrule
\endlastfoot

\multicolumn{3}{@{}l}{\textbf{Tabular-log basics (Sec.~\ref{sec:3.1})}}\\
$X=\{x_0,\dots,x_{N-1}\}$ & sequence & Time-ordered tabular log (entries). \\
$x_n$ & entry & The $n$-th log entry. \\
$t_0<\cdots<t_{N-1}$ & timestamps & Arrival times of entries. \\
$x_n^m$ & attribute value & The $m$-th attribute in entry $x_n$. \\
$M$ & integer & Number of attributes per entry. \\
$\{o_n^0,\dots,o_n^{P-1}\}$ & set & Object attributes extracted from $x_n$. \\
$P$ & integer & Number of object attributes in $x_n$ ($P<M$). \\
$s_n$ & text / node & Event attribute (one per entry; possibly text). \\
$\{f_n^0,\dots,f_n^{Q-1}\}$ & set & Feature attributes extracted from $x_n$. \\
$Q$ & integer & Number of feature attributes in $x_n$ ($Q<M$). \\
$y_n$ & label & Event anomaly label for $x_n$ ($0$ normal, $1$ abnormal). \\
$y_n^p$ & label & Object anomaly label for object $o_n^p$ ($0/1$). \\
$y_n^q$ & label & Feature anomaly label for feature $f_n^q$ ($0/1$). \\

\addlinespace[2pt]
\multicolumn{3}{@{}l}{\textbf{Graphs and dynamics (Sec.~\ref{sec:3.2}--\ref{sec:3.3})}}\\
$\mathcal{G}$ & dynamic graph & Evolving heterogeneous graph over time. \\
$G_n$ & snapshot & Graph snapshot at time $t_n$ (before merging $g_n$). \\
$g_n$ & subgraph & Subgraph constructed from new entry $x_n$. \\
$G_{n+1}\setminus G_n$ & graph diff & Increment between consecutive snapshots; here equal to $g_n$. \\
$\mathcal{V},\mathcal{E}$ & sets & Node and edge sets of the current graph. \\
$v,e$ & node, edge & A node or an edge (generic). \\
$\mathcal{V}_n^o$ & node set & Object nodes appearing in $x_n$. \\
$\mathcal{V}_n^e$ & node set & New event nodes introduced by $x_n$ (events are unique). \\
$\mathcal{E}^+_k$ & edge set & Positive (observed) object–event links in $g_k$. \\
$\mathcal{E}^-_k$ & edge set & Negative samples (non-existent object–event pairs). \\
$\hat{\mathcal{E}}^+_n$ & edge set & Accepted/predicted-positive links at $t_n$. \\
$\hat{\mathcal{V}}^e_n$ & node set & Accepted new events incident to $\hat{\mathcal{E}}^+_n$. \\
$\mathcal{R}_{t_n}$ & set & Per-time link predictions/results at $t_n$. \\
$\{G_n\}_{n=0}^{N-1}$ & sequence & The sequence of snapshots defining $\mathcal{G}$. \\
$G_{t_K},\,G_{t_N}$ & snapshots & Snapshot after train time $t_K$, and final snapshot at $t_N$. \\

\addlinespace[2pt]
\multicolumn{3}{@{}l}{\textbf{Modeling (GNN and scoring; Sec.~\ref{sec:3.4})}}\\
$\theta$ & parameters & Trainable parameters of the GNN. \\
$f_\theta(\cdot)$ & mapping & GNN that computes object-node embeddings on $\mathcal{G}$. \\
$\mathcal{F}$ & encoder & Text (and time-aware) embedding model for events. \\
$h_o,\;h_e$ & vectors & Object and event embeddings, respectively. \\
$\mathrm{reduce}(\cdot,\cdot)$ & operator & Embedding combiner (e.g., concat/diff/dot). \\
$\mathrm{MLP}(\cdot)$ & mapping & Multi-layer perceptron used for scoring. \\
$\sigma(\cdot)$ & function & Sigmoid activation. \\
$s_{o,e}$ & score & Link-normality score for pair $(o,e)$. \\
$\hat{\ell}_{o,e}$ & label & Predicted link label: $\mathbb{1}[s_{o,e}\ge \tau]$. \\
$\mathcal{L}_k$ & loss & Balanced BCE loss at training step $k$. \\
$\eta$ & scalar & Learning rate. \\
$\tau$ & threshold & Operating threshold for prediction. \\
$\rho$ & ratio & Negative sampling ratio. \\

\addlinespace[2pt]
\multicolumn{3}{@{}l}{\textbf{Data splits and indices}}\\
$X_{\text{train}},\,X_{\text{test}}$ & sequences & Training and test splits (chronological). \\
$K$ & integer & Index/time that separates train and test. \\
$N$ & integer & Total number of entries/snapshots. \\
$k,n$ & indices & Training step $k$, evaluation time $n$. \\
$t_k,\;t_n$ & timestamps & Times associated with steps/entries. \\

\end{longtable}

\subsection{Model Design Space}
\label{appendix:model}
We compare two variants in our experiments: (i) Plain (ungated) GAT. We first concatenate the entity-type and entity-ID embeddings and pass them through a feed-forward projection to obtain the initial representation $e_0$. We then run multi-layer, multi-head GATConv on an entity–entity graph induced by shared content to propagate messages and obtain $e_{\text{GAT}}$, which we use as the final entity representation. (ii) Gated fusion. Starting from the same $e_0$ and $e_{\text{GAT}}$, we introduce a global learnable scalar gate $\alpha$ and adaptively combine them via a sigmoid: $e=(1-\sigma(\alpha)),e_0+\sigma(\alpha),e_{\text{GAT}}$. This biases toward $e_0$ when the given signal is weak (or absent) and toward $e_{\text{GAT}}$ when the signal is strong. Both variants share the same link-prediction head: we take the entity representation and the content representation (text and time embeddings concatenated and then projected), compute their element-wise difference, and feed it to an MLP to output the link probability

\subsection{Additional Visualization}
\begin{figure*}[htbp]
  \includegraphics[width=\textwidth]{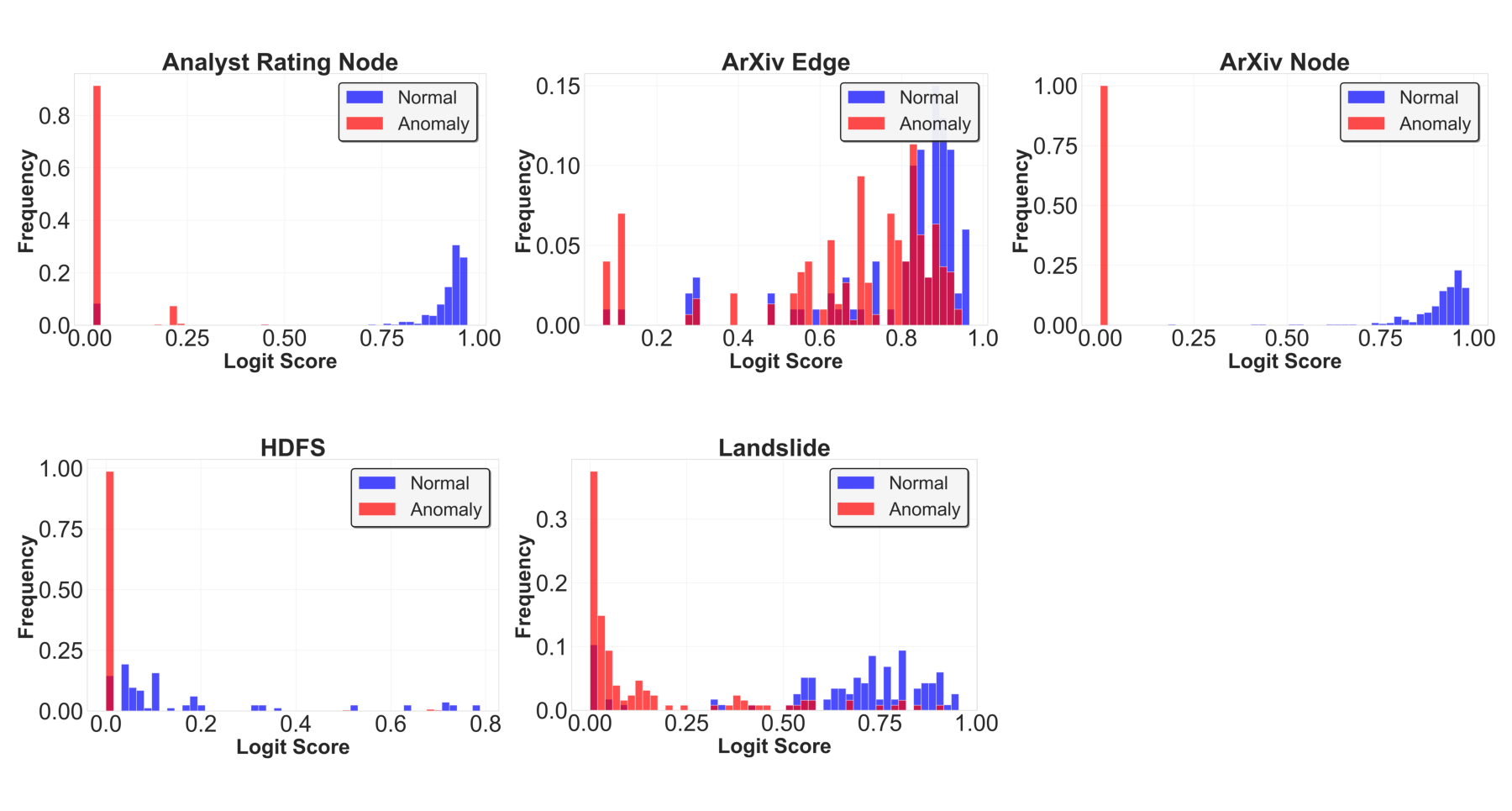}
  \caption{\textbf{Anomaly score distribution of five tasks by GraphLogDebugger.} Score distributions of anomalies and normal examples separate for simpler tasks and mix up for more difficult tasks.}
  \label{fig:vis}
\end{figure*}

Figure~\ref{fig:vis} visualizes the distribution of anomaly likelihood scores of our five evaluation tasks. The score distribution corroborates the main result in Table~\ref{tab:res-detection-efficiency}, that Analyst, Arxiv (Node), and HDFS are three tasks relatively easy, with the score distribution of anomalies and normal examples separate clearly. By contrast, the score of anomalies and normal examples mix up in Arxiv (Edge) and Landslide, indicating that these datasets are more difficult.
\label{appendix:vis}

\subsection{Prompts in RAG}

We list the prompt we used in our RAG baseline as follows:

\begin{lstlisting}[language=Python, caption=Prompt: Arxiv]
"""Build context for ArXiv dataset (authors and paper titles)."""
context = """You are an expert at analyzing author-paper relationships in academic research.

DATASET CONTEXT: This is a dataset of academic papers with their authors and titles.
- Entities (authors): Research authors who wrote the papers
- Content (titles): The titles of the academic papers
- Edge: A connection between an author and a paper title (indicating the author contributed to that paper)

TASK: Determine if the specific author-paper connection (edge) should exist based on historical patterns.

EDGE ANALYSIS TARGET:
"""

context += f"Author: {entity_name}\n"
context += f"Paper Title: {content_name}\n\n"

if similar_contents:
    context += "SIMILAR PAPERS AND THEIR AUTHORS (for reference):\n"
    context += "Use these examples to understand what types of authors typically work on similar papers.\n\n"
    
    for i, content_record in enumerate(similar_contents[:10]):
        content = content_record.get('content', '')
        entities = content_record.get('related_entities', [])
        similarity = content_record.get('similarity', 0.0)
        num_records = content_record.get('num_records', 0)
        
        context += f"{i+1}. Paper Title: {content} (Similarity: {similarity:.3f}, {num_records} records)\n"
        context += f"   Authors who worked on this paper: {', '.join(entities) if entities else 'None'}\n\n"
else:
    context += "No similar papers found in historical data.\n\n"

context += """ANALYSIS QUESTION:
Based on the similar papers and their author patterns, should the specified author-paper connection exist?

EVALUATION CRITERIA:
1. Research Domain Match: Does the author's expertise align with the paper's topic?
2. Historical Patterns: Do authors with similar expertise appear in similar papers?
3. Authorship Likelihood: Is it reasonable that this author would contribute to this type of research?
4. Anomaly Detection: Does this connection seem unusual or out of place compared to patterns in similar papers?

DECISION GUIDELINES:
- edge_exists = True: The author-paper connection makes sense based on research area and historical patterns
- edge_exists = False: The author seems misplaced or unlikely to work on this type of paper (anomalous edge)
- Consider the research fields, methodologies, and typical author patterns shown in similar papers
- An edge is anomalous if the author appears completely unrelated to the research domain of the paper

CONFIDENCE SCORING:
- High confidence (0.8-1.0): Clear patterns in similar papers strongly support/reject the connection
- Medium confidence (0.5-0.7): Some evidence but less certain
- Low confidence (0.0-0.4): Limited historical data or unclear patterns
"""
\end{lstlisting}

\begin{lstlisting}[language=Python, caption=Prompt: HDFS]
"""Build context for HDFS dataset (BlockId focus for detection)."""
context = """You are an expert at analyzing BlockId-log relationships in HDFS distributed file system logs.

DATASET CONTEXT: This is a dataset of HDFS system logs with their Block IDs and log contents.
- Primary Focus: Block IDs (e.g., blk_8215417782549978040, blk_161475555609545016) - unique identifiers for HDFS data blocks
- Content (logs): The actual log messages and operations in the HDFS system that involve specific blocks
- Edge: A connection between a Block ID and a log message (indicating the block is involved in that log operation)

SPECIAL NOTE: For HDFS anomaly detection, we focus specifically on Block ID connections to log messages.
Block IDs should appear BOTH in the BlockId column AND within the log content itself.

TASK: Determine if the specific Block ID-log connection (edge) should exist based on historical patterns.

EDGE ANALYSIS TARGET:
"""

context += f"Block ID: {entity_name}\n"
context += f"Log Content: {content_name}\n"
context += f"Content Analysis: Does '{entity_name}' appear in the log content? {'YES' if entity_name in content_name else 'NO'}\n\n"

if similar_contents:
    context += "SIMILAR LOG MESSAGES AND THEIR BLOCK IDs (for reference):\n"
    context += "Use these examples to understand what types of Block IDs typically appear in similar log messages.\n\n"
    
    for i, content_record in enumerate(similar_contents[:10]):
        content = content_record.get('content', '')
        entities = content_record.get('related_entities', [])
        similarity = content_record.get('similarity', 0.0)
        num_records = content_record.get('num_records', 0)
        
        block_ids = [e for e in entities if e.startswith('blk_')]
        other_entities = [e for e in entities if not e.startswith('blk_')]
        
        context += f"{i+1}. Log Content: {content} (Similarity: {similarity:.3f}, {num_records} records)\n"
        context += f"   Block IDs in this log: {', '.join(block_ids) if block_ids else 'None'}\n"
        if other_entities:
            context += f"   Other entities: {', '.join(other_entities[:3])}{'...' if len(other_entities) > 3 else ''}\n"
        context += "\n"
else:
    context += "No similar log messages found in historical data.\n\n"

context += """ANALYSIS QUESTION:
Based on the similar log messages and their Block ID patterns, should the specified Block ID-log connection exist?

EVALUATION CRITERIA:
1. Block ID Presence: Does the Block ID appear within the log content itself? (This is crucial for HDFS)
2. Log Operation Match: Does the Block ID relate to the HDFS operation described in the log?
3. Historical Patterns: Do similar Block IDs appear in similar log messages?
4. HDFS Block Behavior: Is it reasonable that this Block ID would be involved in this type of operation?
5. Content Consistency: Block ID should be consistent between the BlockId column and the log content

DECISION GUIDELINES:
- edge_exists = True: The Block ID-log connection makes sense based on HDFS block operations and historical patterns
- edge_exists = False: The Block ID seems unrelated to this log message (anomalous edge)
- CRITICAL: If the Block ID does NOT appear in the log content, this is likely anomalous
- Consider HDFS block operations like allocation, storage, replication shown in similar messages
- An edge is anomalous if the Block ID appears completely unrelated to the log operation

CONFIDENCE SCORING:
- High confidence (0.8-1.0): Clear Block ID patterns and content consistency strongly support/reject the connection
- Medium confidence (0.5-0.7): Some evidence but less certain about Block ID relevance
- Low confidence (0.0-0.4): Limited historical data or unclear Block ID patterns

IMPORTANT: Focus specifically on Block ID relationships - Components and Event IDs are secondary for this analysis.
"""
\end{lstlisting}

\begin{lstlisting}[language=Python, caption=Prompt: Analyst and Landslide]
"""Build generic context for unknown datasets."""
context = f"""You are an expert at analyzing entity-content relationships.

EDGE ANALYSIS TARGET:
Entity: {entity_name}
Content: {content_name}

TASK: Determine if this entity-content connection should exist based on historical patterns.
"""

if similar_contents:
    context += "\nSIMILAR EXAMPLES:\n"
    for i, content_record in enumerate(similar_contents[:5]):
        content = content_record.get('content', '')
        entities = content_record.get('related_entities', [])
        context += f"{i+1}. Content: {content}\n   Related entities: {', '.join(entities)}\n\n"

context += """
DECISION: Should this entity-content connection exist?
- edge_exists = True: The connection makes sense based on patterns
- edge_exists = False: The connection seems anomalous
"""
\end{lstlisting}

\end{document}